\documentclass[journal]{IEEEtran}
\usepackage{cite}
\usepackage{amsmath,amssymb,amsfonts}
\usepackage{graphicx}
\usepackage{textcomp}
\usepackage{wrapfig}
\usepackage{epstopdf}

\usepackage{xcolor,soul,framed} 
\usepackage{amssymb}
\colorlet{shadecolor}{yellow}
\usepackage{color,soul}
\usepackage{colortbl}
\soulregister{\cite}7 
\soulregister{\citep}7 
\soulregister{\citet}7 
\soulregister{\ref}7 
\soulregister{\pageref}7 

\usepackage{graphicx}
\graphicspath{{../photo/}}
\DeclareGraphicsExtensions{.pdf,.jpeg,.png,.eps,.bmp}
\usepackage{verbatim}
\usepackage{array}
\usepackage{mdwmath}
\usepackage{mdwtab}
\usepackage{eqparbox}
\usepackage{url}
\usepackage{tabularx}
\usepackage{booktabs}
\usepackage{multirow}
\usepackage{makecell} 
\usepackage{xcolor} 
\usepackage{cite}
\usepackage{diagbox}
\usepackage{float}
\usepackage{balance}
\usepackage{algorithm}
\usepackage{algorithmicx}
\usepackage{algpseudocode}  
\usepackage{soul}

\definecolor{deepgreen}{rgb}{0.0, 0.8, 0.0} 

\def\BibTeX{{\rm B\kern-.05em{\sc i\kern-.025em b}\kern-.08em
    T\kern-.1667em\lower.7ex\hbox{E}\kern-.125emX}}
\definecolor{abstractbg}{rgb}{0.89804,0.94510,0.83137}
\begin{document}

\bstctlcite{IEEEexample:BSTcontrol}
    \title{Towards Aerial Collaborative Stereo: Real-Time Cross-Camera Feature Association and Relative Pose Estimation for UAVs}
  \author{Zhaoying~Wang,
      Wei~Dong
    \thanks{The authors are with the State Key Laboratory of Mechanical System
	and Vibration, School of Mechanical Engineering, Shanghai Jiao Tong University, Shanghai 200240, China (e-mail: wangzhaoying@sjtu.edu.cn; dr.dongwei@sjtu.edu.cn;).}
	\thanks{}
}


\maketitle
\pagestyle{empty}  
\thispagestyle{empty} 


\begin{abstract}

The collaborative visual perception of multiple Unmanned Aerial Vehicles (UAVs) has increasingly become a research hotspot. Compared to a single UAV equipped with a short-baseline stereo camera, multi-UAV collaborative vision offers a wide and variable baseline, providing potential benefits in flexible and large-scale depth perception. In this paper, we propose the concept of a collaborative stereo camera, where the left and right cameras are mounted on two UAVs that share an overlapping FOV. Considering the dynamic flight of two UAVs in the real world, the FOV and relative pose of the left and right cameras are continuously changing. Compared to fixed-baseline stereo cameras, this aerial collaborative stereo system introduces two challenges, which are highly real-time requirements for dynamic cross-camera stereo feature association and relative pose estimation of left and right cameras. To address these challenges, we first propose a real-time dual-channel feature association algorithm with a guidance-prediction structure. Then, we propose a Relative Multi-State Constrained Kalman Filter (Rel-MSCKF) algorithm to estimate the relative pose by fusing co-visual features and UAVs' visual-inertial odometry (VIO). Extensive experiments are performed on the popular onboard computer NVIDIA NX. Results on the resource-constrained platform show that the real-time performance of the dual-channel feature association is significantly superior to traditional methods. Additionally, For the relative pose estimation of two cameras, the adapted multi-UAV Rel-MSCKF algorithm also outperforms the current Pose-Graph Optimization (PGO) manner in real-time performance. The convergence of Rel-MSCKF is assessed under different initial baseline errors. Detailed modeling and analysis are performed for different spatial configurations, including baseline length, view angles, and scene depth. Additionally, the system's robustness is tested against challenges such as asynchronous image acquisition, communication interruptions, and field-of-view occlusions. In the end, we present a potential application of aerial collaborative stereo for remote mapping obstacles in urban scenarios. We hope this work can serve as a foundational study for more multi-UAV collaborative vision research.

Online video: https://youtu.be/avxMuOf5Qcw

\end{abstract}


\maketitle

\section{Introduction}
\label{sec:introduction}
The collaborative visual perception with multiple UAVs gains growing interest\cite{zhang_agile_2022}. Compared to the limited perception range of a single UAV\cite{ParticlesMaps2024}, a multi-UAV system can integrate multiple-view information, holding advantages in flexible and large-scale mapping. In urban post-disaster rescue scenarios, the Global Navigation Satellite System (GNSS) may be restricted due to signal obstruction. The UAV fleet needs to allocate tasks, and collaboratively map and navigate through unknown towns\cite{Multi_UAV_UGV_RL_2021, MultiViewStereo2021}. Compared to the limited depth perception of a single stereo camera, the wide-baseline stereo by different UAVs can collaboratively map remote obstacles ahead with large view disparities, serving for navigation in advance. The foundational modules for high-level collaborative mapping are multi-view visual association and multi-camera relative pose estimation. This paper primarily focuses on the challenges with these two modules in dynamically flying UAVs. In the aerial collaborative stereo, the field-of-view (FOV) and relative pose between left and right cameras are continuously changing. The first challenge is the real-time requirement of cross-camera feature association from large view disparities\cite{sarlin2020superglue}. The second challenge is real-time and continuous relative pose estimation between UAVs during dynamic flight\cite{karrer2018collaborative}.

The ETH ASL\cite{achtelik2011collaborative} firstly explores collaborative stereo and analyzes the system observability using common features to estimate relative pose in numerical simulation. Furthermore, study\cite{karrer2018collaborative} utilizes time delay to simulate two UAVs on the public dataset for relative pose estimation. In their setup, the common features are extracted by ORB descriptor\cite{rublee2011orb} under small view disparities. Study\cite{karrer2021distributed} further simulates a collaborative stereo with two UAVs with downward cameras. Since the two camera views are stable in simulation, the BRISK\cite{BRISK2011} are used for common features. Considering the large view disparities and dynamically changing FOV of stereo setup in real world, high-quality common feature association is desired. The study\cite{sarlin2020superglue} thoroughly evaluates the well-known classic and learned feature association algorithms in Scannet. The compared feature descriptors include ORB\cite{rublee2011orb}, SIFT, SURF\cite{bay2006surf} and SuperPoint\cite{detone2018superpoint}. The compared matching method includes NN matcher and various outlier rejectors: RANSAC\cite{derpanis2010RANSACoverview} as well as learned matcher PointCN\cite{yi2018learning}, SuperGlue\cite{sarlin2020superglue}. SuperPoint, a deep learning-based descriptor, utilizes convolutional neural networks (CNNs) with homographic adaptation to enhance feature detection and description, excelling in complex scenes and varying viewpoints. When combined with SuperGlue, which uses graph neural networks to model feature point relationships, the matching accuracy is significantly improved. This combination effectively addresses challenges such as occlusion and variability\cite{sarlin2020superglue}, making it especially valuable for handling the dynamic viewpoints of different UAVs. Due to the high accuracy demand under large view disparities, omni-swarm\cite{xu2022omni} utilizes SuperPoint\cite{detone2018superpoint} to extract common features cross UAVs and perform Perspective-n-Point (PnP) to calculate the relative pose directly. However, the time-consuming feature detection of SuperPoint causes low-rate feature association. The real-time performance of the above detectors has been compared in \cite{bojanic2019comparison}. A study by \cite{xu2023airvo} shows that SuperPoint and SuperGlue cannot achieve real-time performance on the onboard computer, even with CUDA and TensorRT acceleration. Both high-quality and real-time cross-camera feature association remain challenging in collaborative stereo.

As for the relative pose estimation, direct PnP\cite{xu2022omni} with RANSAC is fast but requires rigorous outlier rejections, which usually leads to noisy and discontinuous results. In order to obtain smooth and accurate relative pose estimation, Kimera-multi\cite{tian2022kimera}, CCM-SLAM\cite{schmuck2019ccm} and COVINS-G\cite{patel_covins-g_2023} first execute two-view PnP to estimate the initial relative pose, then executing global Pose-Grpah-Optimization (PGO) to refine multi-robot trajectories. The PGO process is normally time-consuming and usually executed in a periodic manner. Thus, the relative pose estimation has poor real-time performance. Generally, the run-time of PGO is acceptable in non-real-time tasks, such as multi-robot trajectories alignment and global map merging. However, the highly real-time estimation is essential for the tightly collaborative flying stereo. A real-time and lightweight estimation framework is desired on the resource-constrained onboard computer.

In this paper, to deal with the first real-time feature association challenge, we propose a dual-channel feature association algorithm. The SuperPoint and SuperGLue periodically associate cross-camera features in the guidance channel, then, the guided features are synchronized to the prediction channel for real-time feature prediction. For the second relative pose estimation challenge, we are inspired by MSCKF\cite{mourikis2007multi}, which is a lightweight sliding window filter for single visual-inertial odometry. We re-design the system and propose an adapted MSCKF for the multi-UAV relative pose estimation, called Rel-MSCKF. Specifically, the common features associated with the dual-channel algorithm are effectively fused with local odometry from two UAVs for the relative pose estimation. Benefiting from the real-time and high-quality feature association in the front-end, the Rel-MSCKF can achieve real-time and accurate relative pose estimation between UAVs. We conclude the main contributions of this work as follows:

\begin{itemize}
\item Towards real-time cross-camera feature association in collaborative stereo, a dual-channel algorithm with periodic guidance and real-time prediction is proposed in the front-end. Compared to the single-channel mode with time-consuming SuperPoint and SuperGlue, the guidance channel periodically fuses and synchronizes associated features from SuperPoint and SuperGlue into the high-rate prediction channel. The combination design of guidance and prediction effectively improves the rate of feature association under large view disparities with the resource-constrained onboard computer. Extensive real-world experiments validate the high-quality and real-time superiority of the dual-channel algorithm compared to the existing approaches. 
\item To achieve real-time relative pose estimation between two UAVs in collaborative stereo, a lightweight Rel-MSCKF estimator is developed in the back-end by fusing local odometry with the measurements of common features. The cloned relative pose is predicted by the increments of local odometry from two UAVs and effectively updated by the reprojection of common features. When two UAVs continuously share an overlapping view, the Rel-MSCKF estimator outperforms the PGO-based estimator in terms of real-time and accuracy performance. Further experiments validate that the proposed dual-channel front-end enhances the accuracy performance of the Rel-MSCKF estimator compared with the single-channel front-end.

\end{itemize}

This paper is structured as follows: In Section II, we introduce the dual-channel feature association in the front-end. In Section III, we present the Rel-MSCKF relative pose estimator in the back-end. The experiments are demonstrated in Section IV and the conclusion is finished in Section V.

\section{Dual-channel Feature Association}
\subsection{Method Setup}
The schema diagram of dual-channel feature association is presented in Fig.1. We take UAV $i$ and UAV $j$ to illustrate the cross-camera feature association algorithm. In order to provide a distributed computing architecture between two UAVs, the raw images are processed in the local onboard computer to extract SuperPoint features individually. The UAV $j$ continuously detects and broadcasts the SuperPoint key points to UAV $i$. The UAV $i$ leverages the locally detected SuperPoint key points with the received key points to perform SuperGlue matching. The above process belongs to the guidance channel, aiming to provide high-quality feature matches for the prediction channel. For the prediction channel, the received guided matches are firstly synchronized and stored in the matches club. When the new images arrive, the matches in the club are fastly predicted by Lucas-Kanade optical flow (LK-flow) to the current timestamp. Generally, the LK-flow has the advantages of tracking features on consecutive image streams and fast computation speed with GPU acceleration. In addition, we establish an image caching database ImgDB serving for the image synchronization between the guidance channel and the prediction channel. The workflow in the initialization and update stages will be elaborated in detail below.
\begin{figure}[h]
    \centering
    \includegraphics[width=1.0\linewidth]{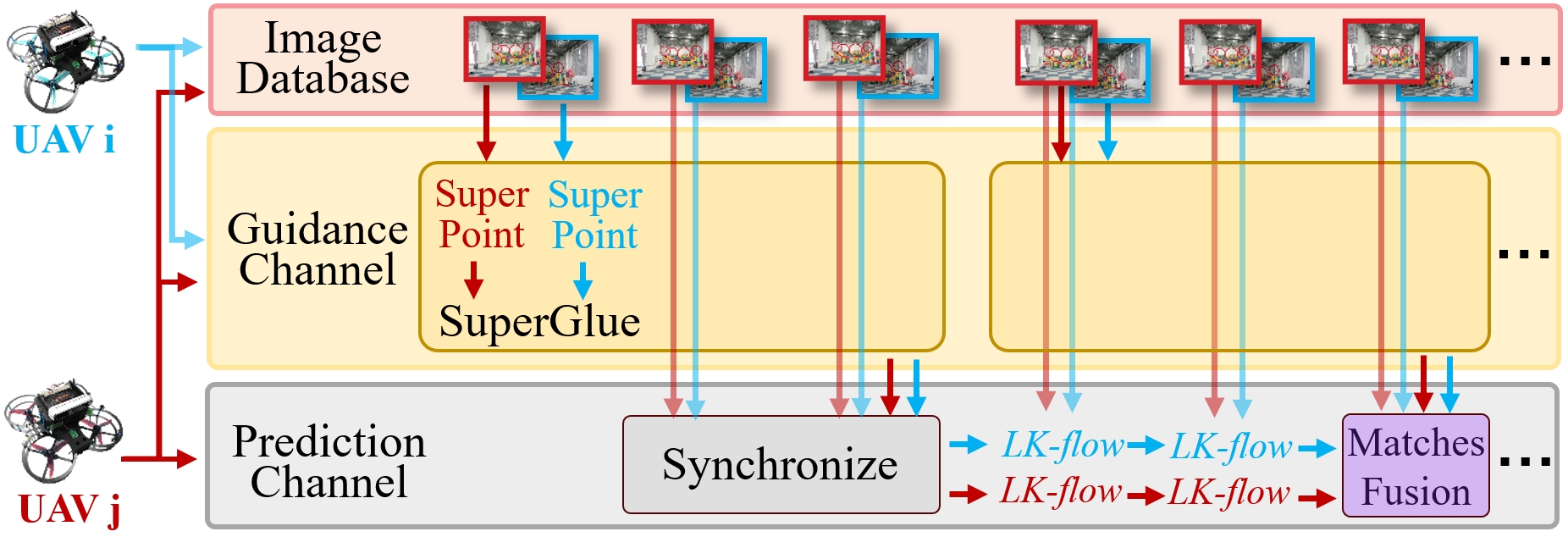}
    \caption{The schema diagram of dual-channel feature association.}
    \label{fig:2}
\end{figure}

\subsection{Initialization Stage} 
The workflow of the initialization stage is presented in Fig.2. Image pairs arrive in sequence at timestamp $t_{1}, t_{2},...,t_{6}$. At the beginning of the algorithm, the guidance channel receives the image pairs $(I_{i}^{1}, I_{j}^{1})$ at $t_{1}$. The SuperPoint descriptor extraction and SuperGlue feature matching require several time steps to complete. The time cost in guidance processing varies according to different devices. Here we take the popular NVIDIA NX Xavier computer\cite{xu2022d2slam} for example illustration. The guidance processing costs every three time steps with 30 Hz image stream. Therefore, the guidance channel generates feature match result after $t_{3}$. The guided feature matches are defined as ${}^{G_{1}}\boldsymbol{f}_{ij}^{1}$, where ${}^{G_{1}}(\cdot)$ means that the guided feature matches are originally generated from image pairs $(I_{i}^{1}, I_{j}^{1})$. $\boldsymbol{f}_{ij}$ means the matched feature points from UAV $i$ and UAV $j$. $(\cdot)^{1}$ means the matches describing the matching condition of timestamp $t_{1}$. Since the matches ${}^{G_{1}}\boldsymbol{f}_{ij}^{1}$ lag behind two frames compared to the current moment, ${}^{G_{1}}\boldsymbol{f}_{ij}^{1}$ will be synchronized first. With the historical image pairs stored in ImgDB, the feature points $({}^{G_{1}}\boldsymbol{f}_{i}^{1}, {}^{G_{1}}\boldsymbol{f}_{j}^{1})$ in ${}^{G_{1}}\boldsymbol{f}_{ij}^{1}$ are iteratively predicted by LK-flow (called Fast-flow) with $(I_{i}^{2}, I_{j}^{2})$ and $(I_{i}^{3}, I_{j}^{3})$ to generate ${}^{G_{1}}\boldsymbol{f}_{ij}^{3}$. Specifically, the LK-flow is performed with feature points on their own image streams. When $(I_{i}^{4}, I_{j}^{4})$ arrive, the prediction channel fastly predicts $\boldsymbol{f}_{ij}^{3}$ to $\boldsymbol{f}_{ij}^{4}$ by LK-flow (called Normal-flow) in frame rate. Meanwhile, the guidance channel acquires $(I_{i}^{4}, I_{j}^{4})$ to start the next process. During the second guidance process, the prediction channel predicts the real-time matches in $t_5$ and $t_6$ by LK-flow.

The Fast-flow synchronizes the outdated guidance matches to the current timestamp before transferring them to the prediction channel. The Normal-flow in the prediction channel effectively utilizes the skipped images to continuously generate feature matches during the second SuperPoint and SuperGlue period. The workflow for initialization is summarized in Algorithm 1. In general, the guidance channel requires three time steps to complete the first guided matches between $t_{k-1}$ and $t_{k}$, which correspond to $t_{3}$ and $t_{4}$ in the above case.

\begin{figure}[h]
    \centering
    \includegraphics[width=1.0\linewidth]{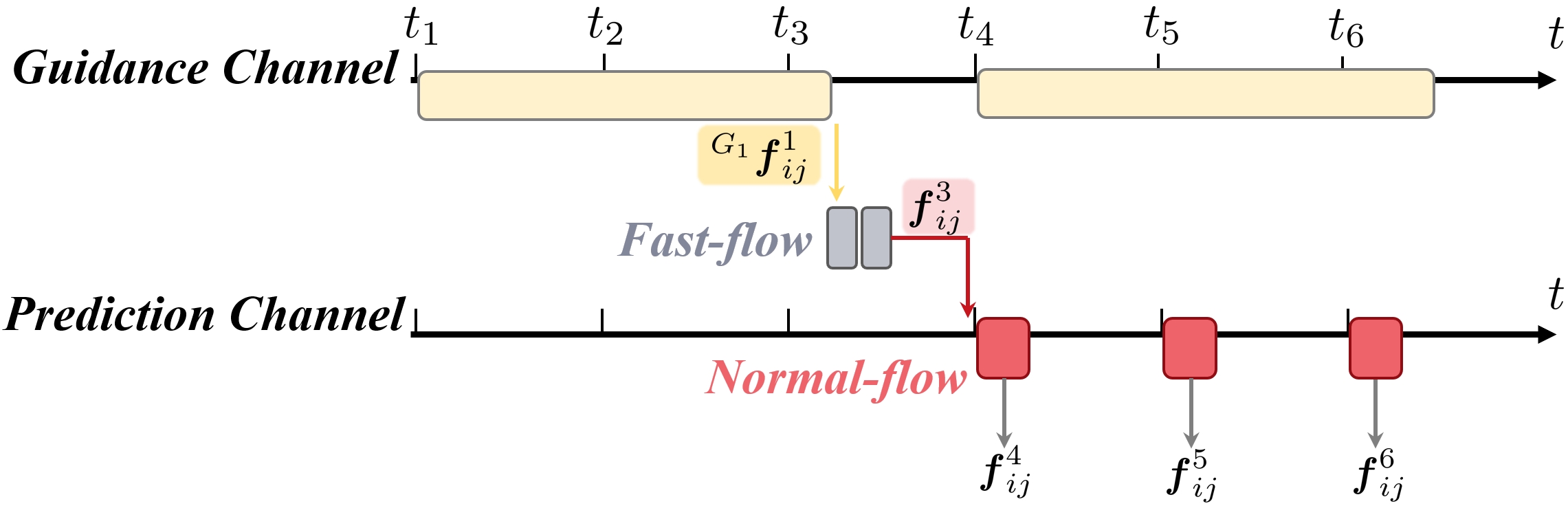}
    \caption{The initialization workflow of dual-channel feature association.}
    \label{fig:3}
\end{figure}

\begin{algorithm}
    \renewcommand{\algorithmicrequire}{\textbf{Input:}}
    \renewcommand{\algorithmicensure}{\textbf{Output:}}
    \caption{Dual-Channel Initialization}
    \label{guidance-prediction feature Initialization}
    \begin{algorithmic}[1]  
        \Require The image pair database ImgDB from $t_1$ to $t_k$
        \Ensure  The feature matches $\boldsymbol{f}_{ij}^{k}$ in current time
        \State Suppose the guidance channel completing the first guided matches in [$t_{k-1}, t_{k}$) ($k > 1$)
        \While {$t <= t_{k}$}
            \State ${}^{G_{1}}\boldsymbol{f}_{i}^{1}, {}^{G_{1}}\boldsymbol{f}_{j}^{1}$ $\gets$ \textbf{SuperPoint}($ I_i^{1}, I_j^{1}$)
            \State $ {}^{G_{1}}\boldsymbol{f}_{ij}^{1}$ $\gets$ \textbf{SuperGlue}(${}^{G_{1}}\boldsymbol{f}_{i}^{1}, {}^{G_{1}}\boldsymbol{f}_{j}^{1}$)
            \State ImgDB $\gets$ $(I_{i}^{t}, I_{j}^{t})$
        \EndWhile
        \For {$n$ in $[1, k-2]$}
            \State ${}^{G_{1}}\boldsymbol{f}_{ij}^{n+1}$ $\gets$ \textbf{Fast-flow}(${}^{G_{1}}\boldsymbol{f}_{ij}^{n}$)
            \State $n = n + 1$
        \EndFor
        \State $\boldsymbol{f}_{ij}^{k}$ $\gets$ \textbf{Normal-flow}(${}^{G_{1}}\boldsymbol{f}_{ij}^{k-1}$)
    \end{algorithmic}
\end{algorithm}

\subsection{Update Stage} 
This part introduces the update stage. With the overlapping views moving, the matched feature points are gradually predicted out of the camera views of $I_i^{k}$ or $I_j^{k}$ by LK-flow. In order to supplement enough matches, the main task of the update stage is to fuse new guided matches from the guidance channel into the prediction channel. The workflow is shown in Fig.3. The guidance channel operates with every three time steps in this case. The second guided matches ${}^{G_{4}}\boldsymbol{f}_{ij}^{4}$ are processed from $(I_{i}^{4}, I_{j}^{4})$ and output the result after $t_6$. Let us focus on the time interval between $t_6$ and $t_7$. The second guided matches ${}^{G_{4}}\boldsymbol{f}_{ij}^{4}$ will fuse with the current matches $\boldsymbol{f}_{ij}^{6}$ in matches club. First, Fast-flow predicts ${}^{G_{4}}\boldsymbol{f}_{ij}^{4}$ to ${}^{G_{4}}\boldsymbol{f}_{ij}^{6}$ for time synchronization. Next, we execute a match fusion of ${}^{G_{4}}\boldsymbol{f}_{ij}^{6}$ and $\boldsymbol{f}_{ij}^{6}$ to obtain $\boldsymbol{\hat{f}}^{6}_{ij}$. After match fusion, the fused matches $\boldsymbol{\hat{f}}^{6}_{ij}$, the Normal-flow continues to predict $\boldsymbol{f}^{7}_{ij}$, $\boldsymbol{f}^{8}_{ij}$, $\boldsymbol{f}^{9}_{ij}$ with coming image frames in real-time.

\begin{figure}[h]
    \centering
    \includegraphics[width=1.0\linewidth]{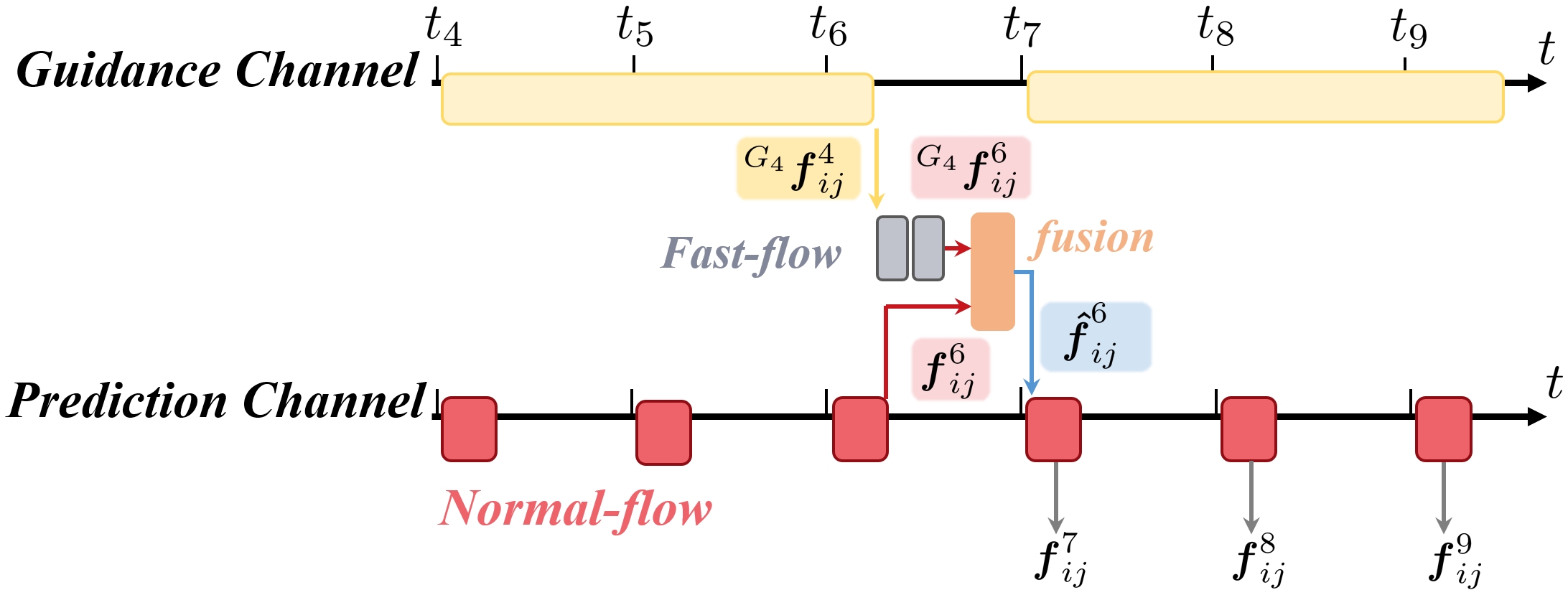}
    \caption{The update workflow of dual-channel feature association.}
    \label{fig:4}
\end{figure}

\begin{figure}[h]
    \centering
    \includegraphics[width=1.0\linewidth]{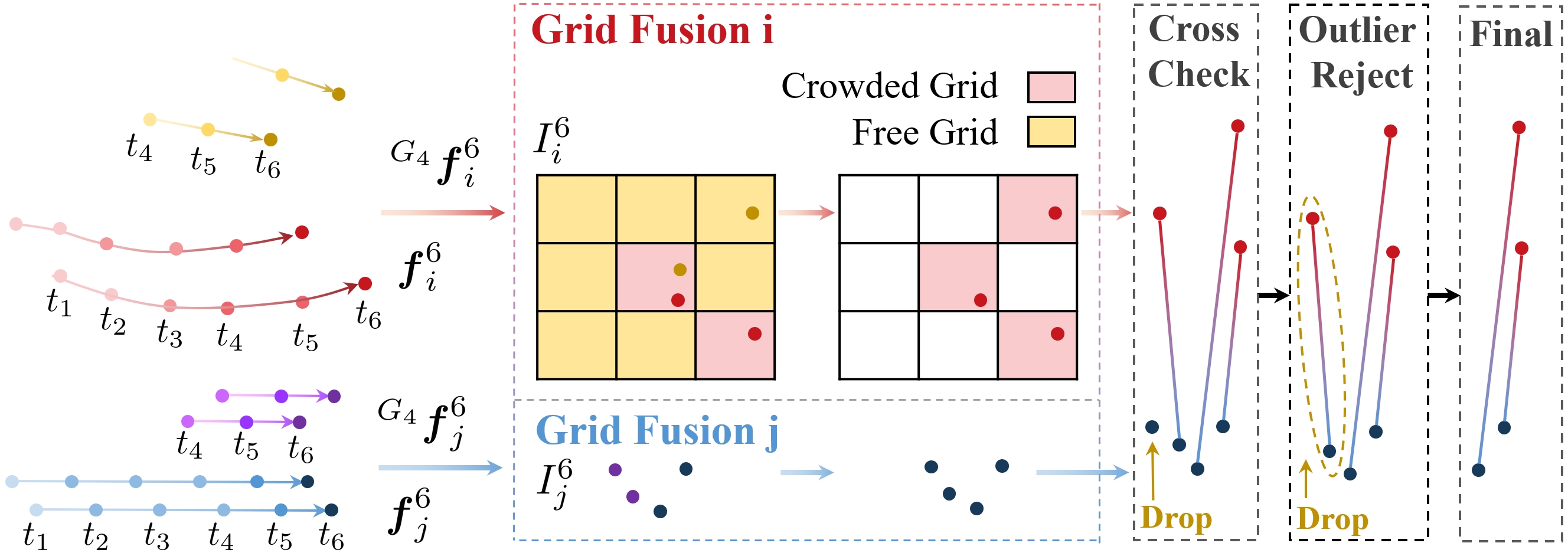}
    \caption{The workflow of match fusion. ${}^{G_{4}}\boldsymbol{f}_{i}^{6}$ and ${}^{G_{4}}\boldsymbol{f}_{j}^{6}$, which are generated from the second guidance, experience twice LK-flow predictions. $\boldsymbol{f}_{i}^{6}$ and $\boldsymbol{f}_{j}^{6}$, which are generated from the first guidance, experience five times of LK-flow predictions.
    }
\end{figure}

\begin{figure}[]
  \centering
  \includegraphics[width=1.0\linewidth]{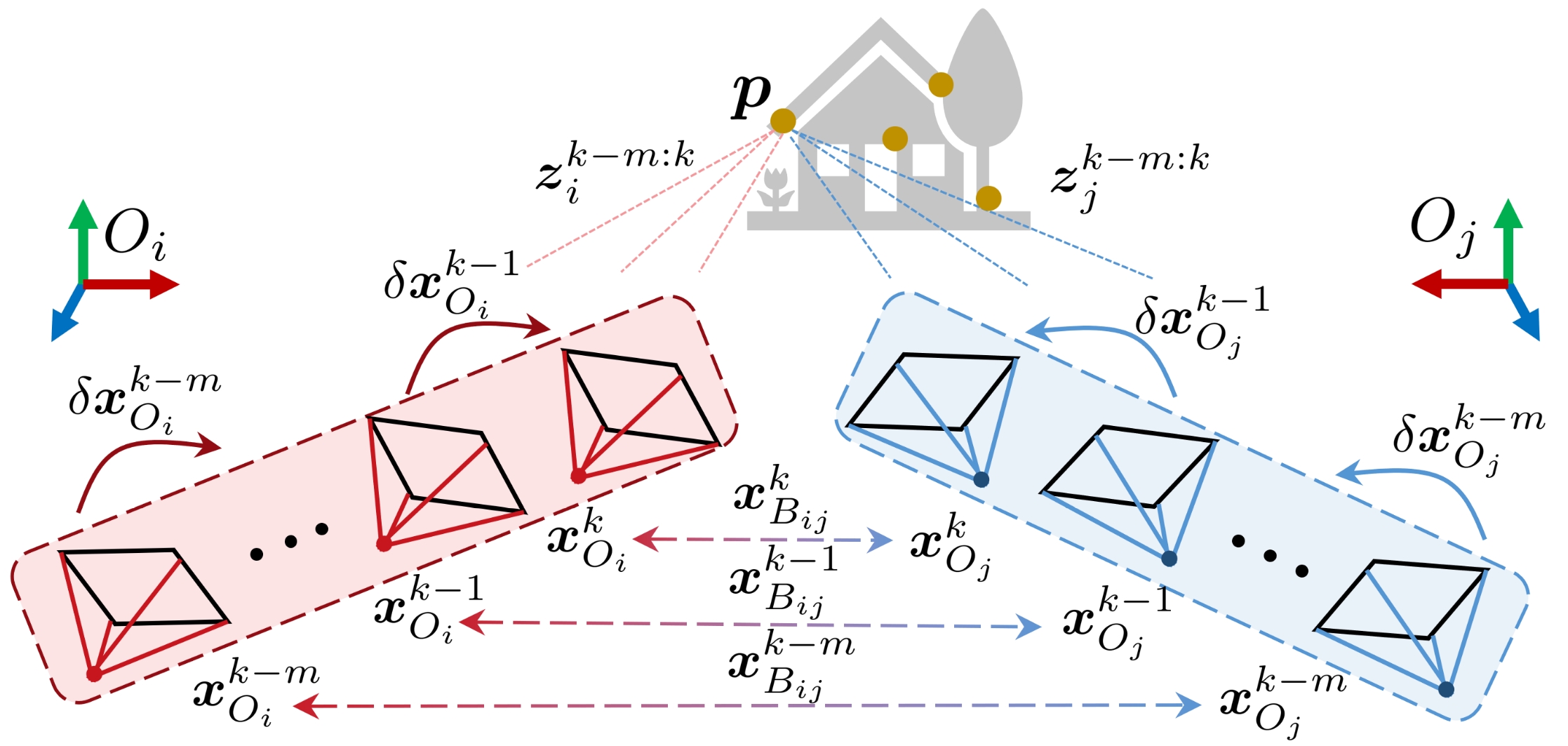}
  \caption{The Rel-MSCKF relative pose estimator in the back-end.}
  \label{fig:5}
\end{figure}

\begin{algorithm}
    \renewcommand{\algorithmicrequire}{\textbf{Input:}}
    \renewcommand{\algorithmicensure}{\textbf{Output:}}
    \caption{Dual-Channel Update}
    \label{guidance-prediction Update}
    \begin{algorithmic}[1]  
        \Require The image pair database ImgDB from $t_{k-m}$ to $t_{k}$, the latest feature match guidance ${}^{G_{k-m}}\boldsymbol{f}_{ij}^{k-m}$, the existing feature match $\boldsymbol{f}_{ij}^{k-1}$ in prediction channel. $k - m$ is the starting time of new guidance procedure.
        \Ensure  The fused feature matches $\boldsymbol{\hat{f}}_{ij}^{k}$
    \State Suppose the guidance channel providing the guided matches ${}^{G_{k-m}}\boldsymbol{f}_{ij}^{k-m}$ in [$t_{k-1},t_{k}$) ($k > 1, m < k$) 
    \For {$n$ in $[k-m, k-2]$}
        \State ${}^{G_{k-m}}\boldsymbol{f}_{ij}^{n+1}$ $\gets$ \textbf{Fast-flow}(${}^{G_{k-m}}\boldsymbol{f}_{ij}^{n}$)
        \State $n = n + 1$
    \EndFor
    \State $\boldsymbol{\hat{f}}_{ij}^{k-1}$ $\gets$ \textbf{MatchFusion}(${}^{G_{k-m}}\boldsymbol{f}_{ij}^{k-1}$, $\boldsymbol{f}_{ij}^{k-1}$)
    \State $\boldsymbol{\hat{f}}_{ij}^{k-1}$ $\gets$ \textbf{CrossCheckAndOutlierReject}($\boldsymbol{\hat{f}}_{ij}^{k-1}$)
    \State $\boldsymbol{\hat{f}}_{ij}^{k}$ $\gets$ \textbf{Normal-flow}($\boldsymbol{\hat{f}}_{ij}^{k-1}$)
    \end{algorithmic}
\end{algorithm}

The match fusion is shown in Fig.4. In order to uniformly distribute the final fused feature points in the image plane, the image $I_{i}^6$ is divided into small grids. The grids occupied by the existing feature points $\boldsymbol{f}_{i}^{6}$ are masked as crowded grids. Only the guided feature points ${}^{G_{4}}\boldsymbol{f}_{i}^{6}$ located in the free grids are allowed to be added. The same grid fusion is executed for the UAV $j$ part. Then a cross-check is performed to delete the isolated points, which are usually generated when their matched points are removed from the crowded grid. Since the feature points may drift after several iterations of LK-flow, we design the outlier rejection module to eliminate the drifting matches. The module includes RANSAC check and directional angle check from the UAV $i$ image to the UAV $j$ image. The general workflow of update stage is concluded as algorithm 2. Suppose the guidance channel providing the guided matches ${}^{G_{k-m}}\boldsymbol{f}_{ij}^{k-m}$ in [$t_{k-1},t_{k}$), which is [$t_{6},t_{7}$) in above condition. The latest feature match guidance is ${}^{G_{k-m}}\boldsymbol{f}_{ij}^{k-m}$, which corresponds to ${}^{G_{4}}\boldsymbol{f}_{ij}^{4}$ as shown above. The existing feature match $\boldsymbol{f}_{ij}^{k-1}$ in prediction channel corresponds to $\boldsymbol{f}_{ij}^{6}$ as depicted above.

\section{Relative Pose Estimation}
\subsection{Method Setup}
Both UAV $i$ and UAV $j$ are equipped with Visual Inertial Odometry (VIO) to estimate pose in its local coordinate. The VIO adopts lightweight FAST corners as front-end, which can be tracked in each frame in the VIO system. While the time-consuming SuperPoint and SuperGlue are executed periodically. To enable any two UAVs to flexibly construct the stereo camera setup within the swarm, the initial coordinate systems between different UAVs are considered unknown in our experiments. As shown in Fig.5, UAV $j$ needs to estimate its relative pose in the body coordinate of UAV $i$. We design a relative pose estimator Rel-MSCKF. Compared to the original MSCKF, the proposed Rel-MSCKF introduces three key differences. First, the estimated states differ: the original MSCKF estimates the pose of a single UAV in the world frame, while Rel-MSCKF estimates the relative pose between two UAVs. Second, the state propagation differs: MSCKF propagates the state and covariance using IMU readings, while Rel-MSCKF propagates based on VIO increments from both UAVs. TThird, the state update differs: MSCKF generates reprojection residuals for its own camera and updates its pose, while Rel-MSCKF calculates reprojection residuals for the neighbor’s camera and updates the relative pose. Therefore, we redesigned the estimated states, cloned states, state propagation, and state update to reconstruct the system. The state vector includes the current relative pose $\boldsymbol{x}_{B_{ij}}^{k}$ and cloned states $\boldsymbol{x}_{B_{ij}}^{k-m:k-1}$.

The relative state vector $\boldsymbol{x}_{B_{ij}}$ from $t_{k-m}$ to $t_{k}$ is decomposed of translation $\boldsymbol{t}_{B_{ij}}$ and orientation $\boldsymbol{q}_{B_{ij}}$ as follows:
\begin{equation}
  \boldsymbol{x}_{B_{ij}} := \left(\boldsymbol{t}_{B_{ij}}^{k-m}, \boldsymbol{q}_{B_{ij}}^{k-m}, ..., \boldsymbol{t}_{B_{ij}}^{k-1},\boldsymbol{q}_{B_{ij}}^{k-1},\boldsymbol{t}_{B_{ij}}^{k},\boldsymbol{q}_{B_{ij}}^{k} \right)
\end{equation}

\subsection{State Propagation}
We first adopt 8-point PnP\cite{hartley1997eightpoint} with the close common features to obtain an initial coarse pose $\boldsymbol{x}^{1}_{B_{ij}}$. Then we start the proposed Rel-MSCKF to continuously estimate the relative pose. Here we illustrate the state propagation process from $t_{k-1}$ to $t_{k}$.

The readings from each VIO from UAV $i$ and UAV $j$ are based on each home coordinates $O_i$ and $O_j$. The increments from $t_{k-1}$ to $t_k$ in the body coordinate are system input $\boldsymbol{u}^{k-1}$, which are presented as follows.

\begin{equation}
\delta\boldsymbol{t}_{B_{i}}^{k-1} = (\boldsymbol{q}_{O_{i}}^{k-1})^{-1} \cdot (\boldsymbol{t}_{O_{i}}^{k} - \boldsymbol{t}_{O_{i}}^{k-1})
\end{equation}
\begin{equation}
\delta\boldsymbol{q}_{B_{i}}^{k-1} = (\boldsymbol{q}_{O_{i}}^{k-1})^{-1} \cdot \boldsymbol{q}_{O_{i}}^{k}
\end{equation}
\begin{equation}
\delta\boldsymbol{t}_{B_{j}}^{k-1} = (\boldsymbol{q}_{O_{j}}^{k-1})^{-1} \cdot (\boldsymbol{t}_{O_{j}}^{k} - \boldsymbol{t}_{O_{j}}^{k-1})
\end{equation}
\begin{equation}
\delta\boldsymbol{q}_{B_{j}}^{k-1} = (\boldsymbol{q}_{O_{j}}^{k-1})^{-1} \cdot \boldsymbol{q}_{O_{j}}^{k}
\end{equation}

With the estimation ($\boldsymbol{t}^{k-1}_{B_{ij}}$, $\boldsymbol{q}^{k-1}_{B_{ij}}$) at $t_{k-1}$ and the increment of VIO ($\delta\boldsymbol{t}_{B_{i}}^{k-1}, \delta\boldsymbol{q}_{B_{i}}^{k-1}, \delta\boldsymbol{t}_{B_{j}}^{k-1}, \delta\boldsymbol{q}_{B_{j}}^{k-1}$), we could construct a transformation loop and predict the relative pose at $t_k$:
\begin{equation}
\hat{\boldsymbol{t}}_{B_{ij}}^{k} = (\delta\boldsymbol{q}_{B_{i}}^{k-1})^{-1} (\boldsymbol{t}_{B_{ij}}^{k-1} + \boldsymbol{q}_{B_{ij}}^{k-1} \cdot \delta\boldsymbol{t}_{B_{j}}^{k-1} - \delta\boldsymbol{t}_{B_{i}}^{k-1})
\end{equation}

\begin{equation}
\hat{\boldsymbol{q}}_{B_{ij}}^{k} = (\delta\boldsymbol{q}_{B_{i}}^{k-1})^{-1} \cdot \boldsymbol{q}_{B_{ij}}^{k-1} \cdot \delta\boldsymbol{q}_{B_{j}}^{k-1}
\end{equation}

The associated covariance propagation from $t_{k-1}$ to $t_{k}$ is as follows:
\begin{equation}
\hat{\boldsymbol{P}}^{k}=\boldsymbol{F}^{k-1} \boldsymbol{P}^{k-1} (\boldsymbol{F}^{k-1})^{T}+\boldsymbol{G}^{k-1} \boldsymbol{Q}^{k-1} (\boldsymbol{G}^{k-1})^{T}
\end{equation}
$\boldsymbol{P}^{k-1}$ is the covariance matrix of the state uncertainty. The covariance matrix $\boldsymbol{Q}^{k-1}$ is the noise from the incremental motion provided by VIO. The matrix $\boldsymbol{F}^{k-1}$ and $\boldsymbol{G}^{k-1}$ are the jacobians of the predicted state with respect to the state variables and the VIO input, respectively:

\begin{eqnarray}
\begin{aligned}
\setlength{\arraycolsep}{1.2pt}
&\boldsymbol{F}^{k-1}:=\frac{\partial \hat{\boldsymbol{x}}^{k}}{\partial \boldsymbol{x}^{k-1}} = 
\\
& \left[\begin{array}{cc}
    (\delta\boldsymbol{q}_{B_{i}}^{k-1})^{-1} & - \boldsymbol{D} \left[\delta\boldsymbol{t}_{B_{j}}^{k-1} \right]_{\times} \\
    \boldsymbol{0}_{3 \times 3} & (\delta\boldsymbol{q}_{B_{i}}^{k-1})^{-1}
    \end{array}\right]
\end{aligned}
\end{eqnarray}

\begin{equation}
\begin{aligned}
\setlength{\arraycolsep}{0.3pt}
&\boldsymbol{G}^{k-1} :=\frac{\partial \hat{\boldsymbol{x}}^{k}}{\partial \boldsymbol{u}^{k-1}} = \\
& \left[\begin{array}{cccc}
    - (\delta\boldsymbol{q}_{B_{i}}^{k-1})^{-1} & \boldsymbol{C} & \boldsymbol{D} & \boldsymbol{0}_{3 \times 3} \\
    \boldsymbol{0}_{3 \times 3} & - \delta\boldsymbol{q}_{B_{i}}^{k-1} & \boldsymbol{0}_{3 \times 3} & \boldsymbol{D}
    \end{array}\right]
\end{aligned}
\end{equation}
where $\boldsymbol{C}$ is:
\begin{equation}
\boldsymbol{C} = - (\delta\boldsymbol{q}_{B_{i}}^{k-1})^{-1}  [\boldsymbol{t}_{B_{ij}}^{k-1} + \boldsymbol{q}_{B_{ij}}^{k-1} \cdot \delta\boldsymbol{t}_{B_{j}}^{k-1} - \delta\boldsymbol{t}_{B_{i}}^{k-1}]_{\times} \delta\boldsymbol{q}_{B_{i}}^{k-1}
\end{equation}
The $\boldsymbol{D}$ represents $(\delta\boldsymbol{q}_{B_{i}}^{k-1})^{-1} \boldsymbol{q}^{k-1}_{B_{ij}}$.

Then we perform state augmentation, where the predicted $(\hat{\boldsymbol{t}}^{k}_{B_{ij}}, \hat{\boldsymbol{q}}^{k}_{B_{ij}})$ are inserted into the state vector and the $\hat{\boldsymbol{P}}^{k}$ is added into full state covariance $\boldsymbol{P}$.

\subsection{State Update}
The 3D position of feature $\boldsymbol{p}_{O_{i}}$ is initialized by fusing 2D observations and depth image from $t_{k-m}$ to $t_k$ by UAV $i$. Then $\boldsymbol{p}_{O_{i}}$ reprojects to the associated poses of UAV $j$ to update the relative poses $(\boldsymbol{t}^{k-m}_{B_{ij}}, \boldsymbol{q}^{k-m}_{B_{ij}},...,\boldsymbol{t}^{k}_{B_{ij}}, \boldsymbol{q}^{k}_{B_{ij}})$. We take $(\boldsymbol{t}^{k}_{B_{ij}}, \boldsymbol{q}^{k}_{B_{ij}})$ to illustrate the observation model. The predicted 2D measurement $\hat{\boldsymbol{z}}^{k}$ in UAV $j$ is generated by the feature point $\boldsymbol{p}_{O_{i}}$ with the unique feature ID. Thus, the predicted 2D features $\hat{\boldsymbol{z}}^{k}$ can be associated with the actual 2D measurements $\tilde{\boldsymbol{z}}^{k}$.
\begin{equation}
  \hat{\boldsymbol{z}}^{k} = \boldsymbol{h}_{dist}(\Pi(\hat{\boldsymbol{p}}^{k}_{B_{j}})+ \zeta_i) + \boldsymbol{n}_{p}
\end{equation}
\begin{equation}
  \hat{\boldsymbol{p}}^{k}_{B_{j}} = \boldsymbol{q}^{k}_{BC} (\hat{\boldsymbol{q}}^{k}_{B_{ij}})^{-1} ((\boldsymbol{q}^{k}_{O_{i}})^{-1}(\boldsymbol{p}_{O_{i}} - \boldsymbol{t}^{k}_{O_{i}}) - \hat{\boldsymbol{t}}^{k}_{B_{ij}}) + \boldsymbol{t}^{k}_{BC}
\end{equation}
where $\boldsymbol{h}_{dist}$ is the image distortion model with parameter $\zeta_i$. $\Pi$ is the pinhole camera model. $(\boldsymbol{q}^{k}_{BC}, \boldsymbol{t}^{k}_{BC})$ is to transfer the body coordinate to the camera coordinate. $\boldsymbol{n}_{p}$ is the zero-mean white Gaussian measurement noise with covariance $\boldsymbol{R}^{k-m:k} = \boldsymbol{\sigma}^{k-m:k}$. 

The real measured pixel location of the feature at $t_k$ is $\tilde{\boldsymbol{z}}^{k}$. Then we can formulate the stacked error residuals from $t_{k-m}$ to $t_k$ as:
\begin{equation}
  \boldsymbol{r}^{k-m:k} = \tilde{\boldsymbol{z}}^{k-m:k} - \hat{\boldsymbol{z}}^{k-m:k}
\end{equation}
The linearization of this measurement model yields the following:
\begin{equation}
  \boldsymbol{r}^{k-m:k} = \boldsymbol{H}_{x}^{k-m:k} \hat{\boldsymbol{x}}_{B_{ij}}^{k-m:k} + \boldsymbol{H}_{f}^{k-m:k} \boldsymbol{p}_{O_{i}} + \boldsymbol{n}_{p}^{k-m:k}
\end{equation}
We temporarily drop $(\cdot)^{k-m:k}$ for simplicity. Then we calculate the left nullspace of $\boldsymbol{H}_{f}$ (i.e., $\boldsymbol{Q}^{\top} \boldsymbol{H}_{f} =\boldsymbol{0}$) using Givens Rotation and apply to the above equation to remove the dependence of features\cite{geneva2020openvins}.
\begin{equation}
   \boldsymbol{Q}^{\top} \boldsymbol{r} = \boldsymbol{Q}^{\top} \boldsymbol{H}_{x} \hat{\boldsymbol{x}}_{B_{ij}} + \boldsymbol{Q}^{\top} \boldsymbol{H}_{f} \boldsymbol{p}_{O_{i}} + \boldsymbol{Q}^{\top} \boldsymbol{n}_{p}
\end{equation}

\begin{equation}
  \Rightarrow \tilde{\boldsymbol{r}} = \tilde{\boldsymbol{H}}_{x} \hat{\boldsymbol{x}}_{B_{ij}} + \tilde{\boldsymbol{n}}_{p}
\end{equation}
Then we execute a Mahalanobis test for the residuals $\tilde{\boldsymbol{r}}^{k-m:k}$. Finally, we can compute the Kalman gain, and update the relative state and covariance:
\begin{equation}
  \boldsymbol{S}^{k-m:k} = \tilde{\boldsymbol{H}}^{k-m:k}_{x} \hat{\boldsymbol{P}}^{k-m:k} (\tilde{\boldsymbol{H}}^{k-m:k}_{x})^{\top} + \boldsymbol{R}^{k-m:k}
\end{equation}
\begin{equation}
  \boldsymbol{K}^{k-m:k} = \hat{\boldsymbol{P}}^{k-m:k} (\tilde{\boldsymbol{H}}^{k-m:k}_{x})^{\top} (\boldsymbol{S}^{k-m:k})^{-1}
\end{equation}
\begin{equation}
  \boldsymbol{x}_{B_{ij}}^{k-m:k} = \hat{\boldsymbol{x}}_{B_{ij}}^{k-m:k} \oplus \boldsymbol{K}^{k-m:k} \tilde{\boldsymbol{r}}^{k-m:k}
\end{equation}
\begin{equation}
  \boldsymbol{U} = \boldsymbol{I} - \boldsymbol{K}^{k-m:k} \tilde{\boldsymbol{H}}^{k-m:k}_{x}
\end{equation}
\begin{equation}
  \boldsymbol{P}^{k-m:k} = \boldsymbol{U} \hat{\boldsymbol{P}}^{k-m:k} \boldsymbol{U}^{\top} + \boldsymbol{K}^{k-m:k} \boldsymbol{R}^{k-m:k} \boldsymbol{K}^{k-m:k}
\end{equation}

\section{Experiment and Discussion}
\subsection{Experiment Setup}
The system architecture is shown in Fig.6. In the onboard sensors part, Xsens-Mti-630 is chosen as IMU, which provides angular velocity and acceleration for VIO propagation. A stereo Intel Realsense D455 camera is placed on the front-top of the UAV, providing monocular color images for common feature detection and VIO. The VIO part adopts OpenVINS\cite{geneva2020openvins} to estimate the local odometry of the UAV. In the front-end, we distribute the computation among UAVs. Each UAV performs feature detection independently and broadcasts the results to other UAVs. One of the UAVs is selected to perform feature matching and combine the prediction channel to obtain feature matches in real-time. Each UAV is equipped with a TP-Link WDR7650 mesh router, which supports high bandwidth (up to 20 MB/s) and low-latency (low than 15 ms) data exchange. The exchanged data includes the current VIO pose, compressed image frame (for making image pairs), and extracted features. Finally, the back-end is deployed on master UAV $i$ to estimate the relative pose with the VIO prediction and update from feature matches. 

When there are more UAVs involved in the collaborative system in the future, we plan to adopt a serial cascading network structure. Each robot possesses two roles, serving as a slave to the former robot while acting as a master to the latter robot. Communication is also established between the former robot and the latter robot. For example, the UAV $j$ can serve as a master to the third UAV $k$. Each former master and latter slave are chosen by the nearest neighbor. To guarantee sufficient overlapping views, UAV $j$ will align its yaw orientation with UAV $i$ in our experiments. The relative position along the baseline is also adjusted according to the environment. We adopt the PID policy to control the yaw and relative position of slave UAV j, where the current relative pose estimation is provided by Rel-MSCKF.

All programs are developed by the Robot Operating System (ROS) in the Linux Ubuntu 20.04 system and coded by C++14. We select NVIDIA Xavier NX (8GB) as the onboard computer, which is widely adopted in UAV swarm\cite{xu2022d2slam}. Four self-constructed typical scenarios and four public datasets are selected for evaluation as shown in Fig.7.

\begin{figure}[]
  \centering
  \includegraphics[width=1.0\linewidth]{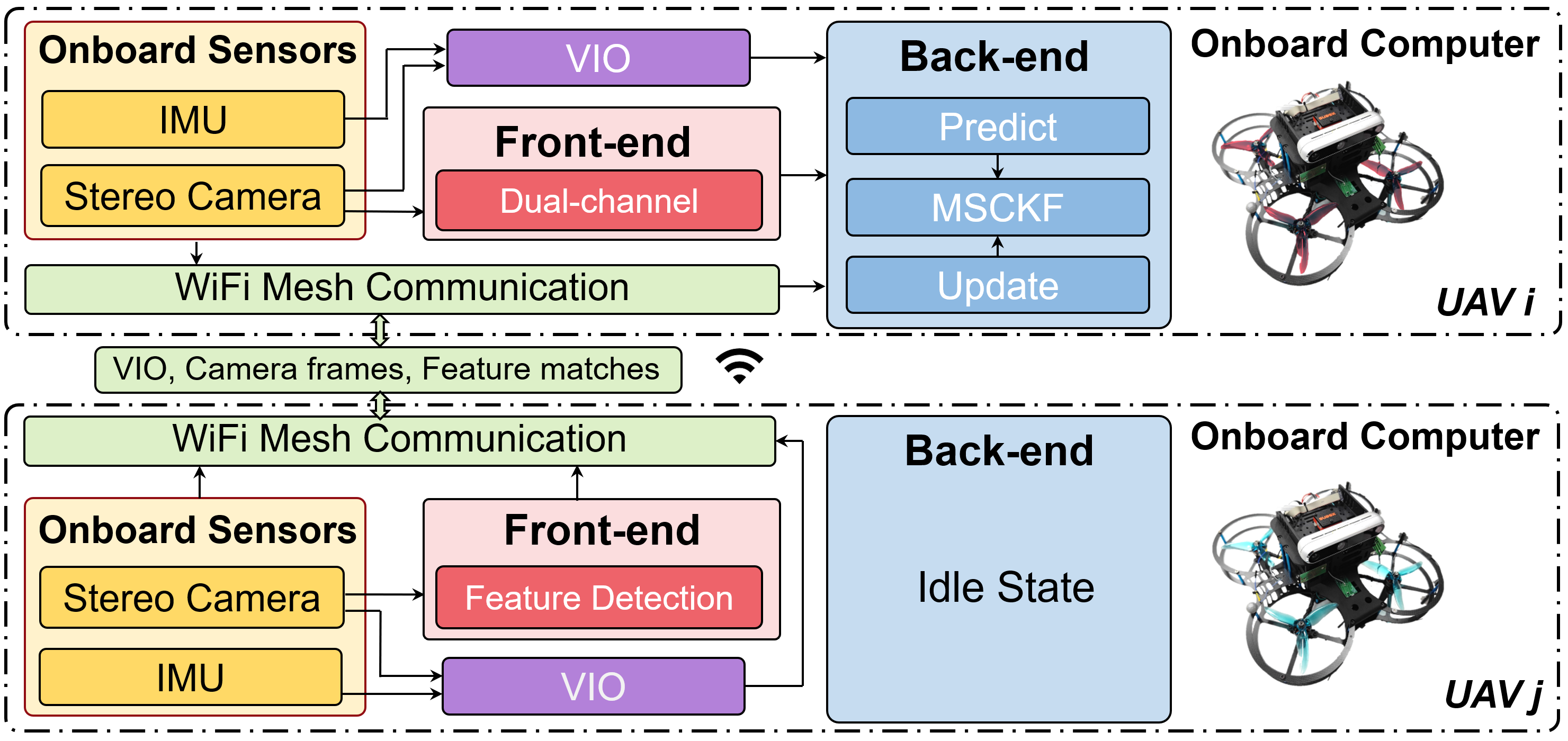}
  \caption{The system architecture of two collaborative UAVs.}
\end{figure}

\begin{figure}[]
  \centering
  \includegraphics[width=1.0\linewidth]{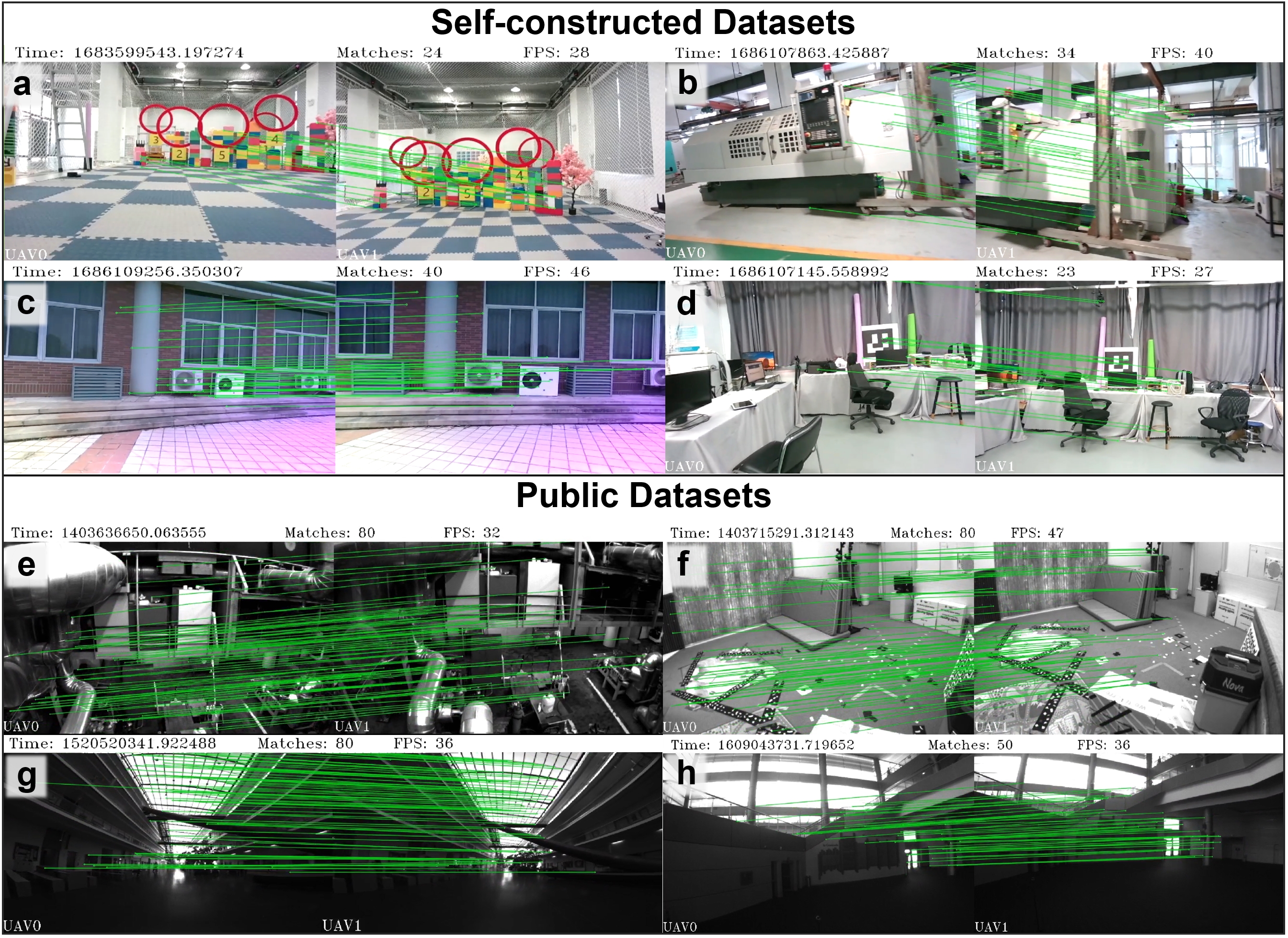}
  \caption{The front-end experimental scenarios. (a) is a flight field with motion capture system. (b) is a factory scenario. (c) is an outdoor building facade. (d) is a laboratory room. (e) is the MH\_01\_easy sequence of EuRoC datasets. (f) is the Vicon\_room\_01 sequence of EuRoC datasets. (g) is the magistrale1 sequence of TUM VI datasets. (h) is the nya\_01 sequence of VIRAL datasets.}
\end{figure}

\subsection{Run-time Analysis}
We compare the run-time of the proposed dual-channel method with other popular feature association methods, including the original SuperPoint with SuperGlue (called SS), ORB with NN match and RANSAC filter (called ORB\_NR), and SURF with NN match and RANSAC filter (called SURF\_NR). ORB is chosen for its rotational and affine invariance, making it a widely used traditional feature extraction method for matching in SLAM tasks. SURF, on the other hand, offers more robust feature extraction by utilizing an efficient approximation of the Hessian matrix for keypoint detection and a distribution-based descriptor for feature matching, albeit at the expense of computational efficiency. In our proposed dual-channel structure, the guidance channel is based on SuperPoint and SuperGlue, and the prediction channel is accelerated by LK-flow. Therefore, our method is also called SSF. In the software configurations, we deploy CUDA 11.4 together with TensorRT 8.4 to accelerate the network inference of SuperPoint and SuperGlue. The LK-flow is implemented by OpenCV with CUDA acceleration. 

The Realsense D455 provides each UAV with a color image stream with 640 $\times$ 480 pixels at 30 Hz. In the two-UAV system, UAV $j$ performs feature detection and broadcasts to UAV $i$. The UAV $i$ performs local feature detection and feature match. Thus the run-time of feature association mainly depends on UAV $i$. For the single-channel SS, SURF\_NR and ORB\_NR approaches, the run-time consists of one feature detection time and one feature match time. For the dual-channel SSF approach, the run-time of feature association is no longer constrained by feature detection and match in the guidance channel. The run-time depends on the prediction channel, which consists of feature normal-flow, fast-flow, and match fusion. Table \uppercase\expandafter{\romannumeral1} shows the detailed run-time statistics on self-constructed datasets. The run-time is the average statistic tested on NVIDIA NX board by 30 times with a maximum of 150 key points per image.

\begin{table}[]
    \centering
    \caption{Run-time Analysis on Self-constructed Datasets}
    \setlength{\tabcolsep}{2.5mm}{
    \begin{tabular}{lllll}
    \toprule
    Modules    & SSF      & SS   & SURF  & ORB  \\ \hline
    Feature Detect         & 37.3 ms    & 37.3 ms & 126.5 ms & 35.3 ms\\
    Feature Match          & 37.0 ms    & 37.0 ms & 31.4 ms & 40.7 ms\\
    Normal-flow       & 7.4 ms     & -    & -     & -    \\
    Fast-flow         & 15.6 ms    & -    & -     & -    \\
    Matches Fusion         & 2.7 ms     & -    & -     & -    \\
    Total (min/max) & \textbf{7.4/25.7} ms & 74.3 ms & 157.9 ms & 76.0 ms\\
    \bottomrule
    \end{tabular}
    }
  \end{table}

  Furthermore, we analyze the detailed run-time along the sequence of image frames in Fig.8. The image frame rate is configured at 30 Hz (33 ms per image). We can see which image frames are processed or skipped and their time cost. With the dual-channel design, SSF can utilize each image frame to generate high-rate feature matches.
  
  \begin{figure}[]
    \centering
    \includegraphics[width=1.0\linewidth]{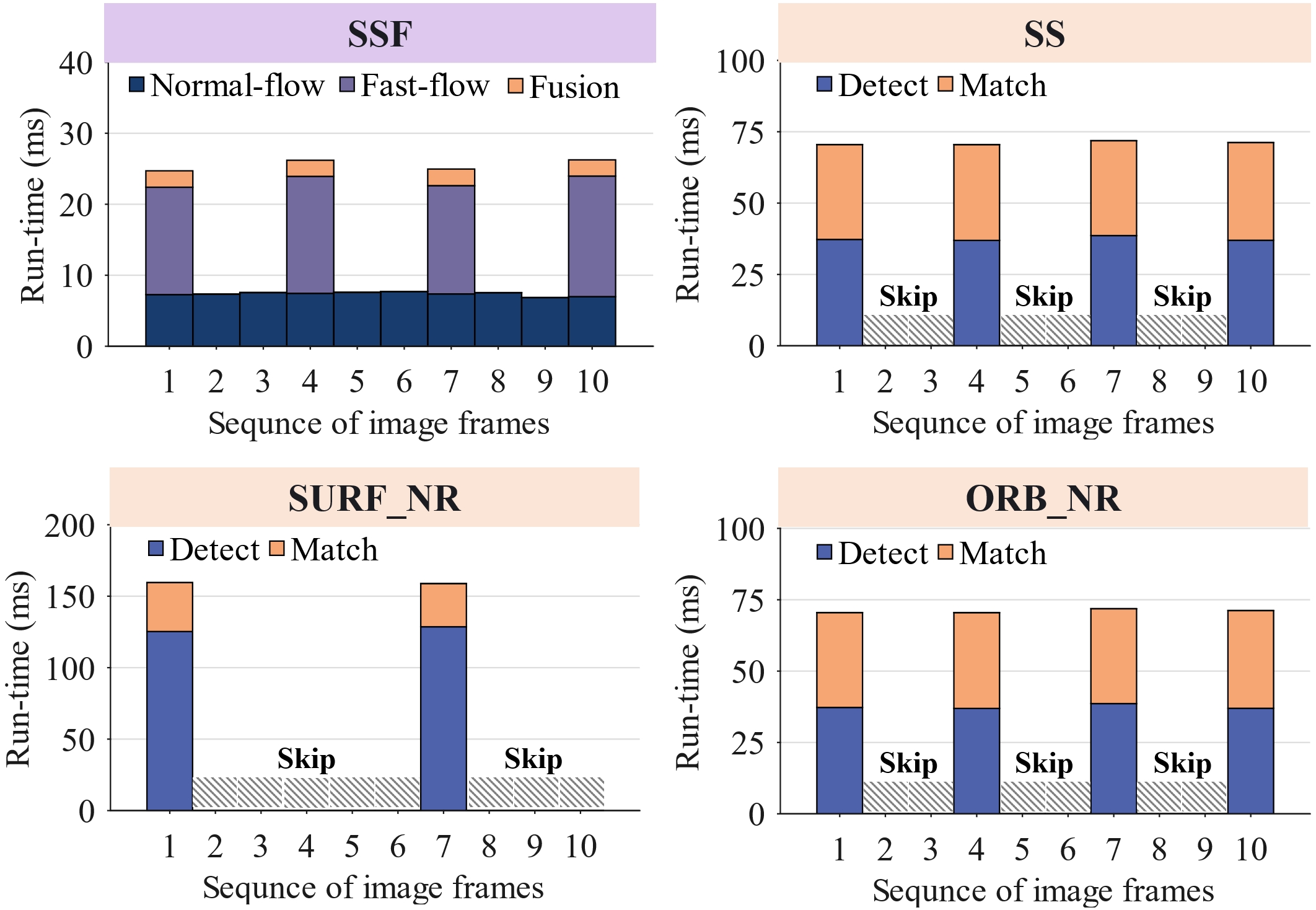}
    \caption{The detailed run-time of SSF, SS, SURF, and ORB along image sequence. SSF periodically receives guided feature matches and executes fast-flow and matches fusion at $t_1,t_4,t_7,t_{10}$. At the guidance interval time stamps, SSF executes normal-flow to predict feature matches with current images. The total run-time of feature detection and match in SS and ORB\_NR is about 75 ms, which will skip the subsequent two frames of images. The execution time of one SURF\_NR is about 150ms, which makes five subsequent frames skipped.}
\end{figure}

We also evaluate the four feature association methods on the widely adopted public datasets, which are EuRoC\cite{burri2016euroc}, TUM VI\cite{schubert2018tumvi}, VIRAL\cite{nguyen2022ntu}. Since there is no existing dataset for multi-UAV with stable overlapping camera views, we generate two UAV datasets based on the single UAV dataset. Specifically, we simulate the two UAVs using one sequence for UAV $i$ and the same one for UAV $j$ but with a time-offset $t_d$. The average run-time statistics are shown in Table {\uppercase\expandafter{\romannumeral2}}. For the experiment on the EuRoC MH\_01\_easy, the SSF has better real-time performance (average 13.1 ms) compared to other methods. The other evaluation sequences on EuRoC, TUM VI and VIRAL show similar results.

\begin{table}[]
\centering
\caption{Run-time Analysis on Public Datasets}
\setlength{\tabcolsep}{1.6mm}{
\begin{tabular}{@{}llllll@{}}
\toprule
Datasets info       & $t_d$      & SSF        & SS       & SURF\_NR       & ORB\_NR   \\ \midrule
EuRoC MH\_01\_easy     & 2s               & \textbf{13.1} ms   & 78.2 ms         & 155.4 ms              & 84.1 ms \\
EuRoC Vicon\_room\_01  & 2s            & \textbf{13.8} ms  & 77.5 ms         & 149.7 ms              & 82.5 ms         \\
TUM VI magistrale1     & 3s            & \textbf{14.1} ms & 78.6 ms        & 144.2 ms          & 84.8 ms        \\
VIRAL nya\_01          & 3s            & \textbf{14.3} ms  & 74.2 ms        & 139.7 ms           & 75.6  ms       \\ \bottomrule
\end{tabular}
}
\end{table}

\subsection{Relative Pose Estimation}
We evaluate the performance of relative pose estimation from four perspectives. First, we assess the convergence of relative pose estimation under inaccurate initial baseline conditions. Second, we validate the correction performance of Rel-MSCKF under VIO drifting conditions. Third, we compare the convergence speed and accuracy of our Rel-MSCKF with different feature association front-ends. Finally, we compare the Rel-MSCKF performance with the optimization-based backend method.

The real-world experiments are performed in the SEIEE flight field and the Meta flight field as shown in Fig.9 (a)(b). Both flight fields are equipped with the Nokov motion capture system to provide ground truth data for quantitative analysis. The Nokov system could provide position precision of 1 mm and an attitude precision of 0.1${}^{\circ}$. The SEIEE flight field has size of 8m $\times$ 4m. The coordinate obeys ENU rules, which is X-front, Y-left, and Z-up. The wall containing the feature landmarks is placed at the depth of 8m. The Meta flight field has smaller size of 6m $\times$ 3.5m. We mainly perform experiments in SEIEE flight filed. Validation using the public VIRAL dataset is shown in Fig.9 (c).

The determination of the baseline is based on three factors. First, geometric calculations are used to maximize the common feature area, a forward-looking view is selected while keeping the baseline as small as possible. Second, to ensure flight stability and minimize interference from downwash airflow between the UAVs, the baseline distance is maintained at a minimum of 2 meters. Third, due to the constraints of the flight field, long baselines (greater than 4 meters) result in a reduced common feature area and may exceed the coverage of the motion capture system. Therefore, the baseline in our system is chosen to range from 2m to 4m, with 3m being a moderate and commonly used value.

\begin{figure}[h]
  \centering
  \includegraphics[width=1.0\linewidth]{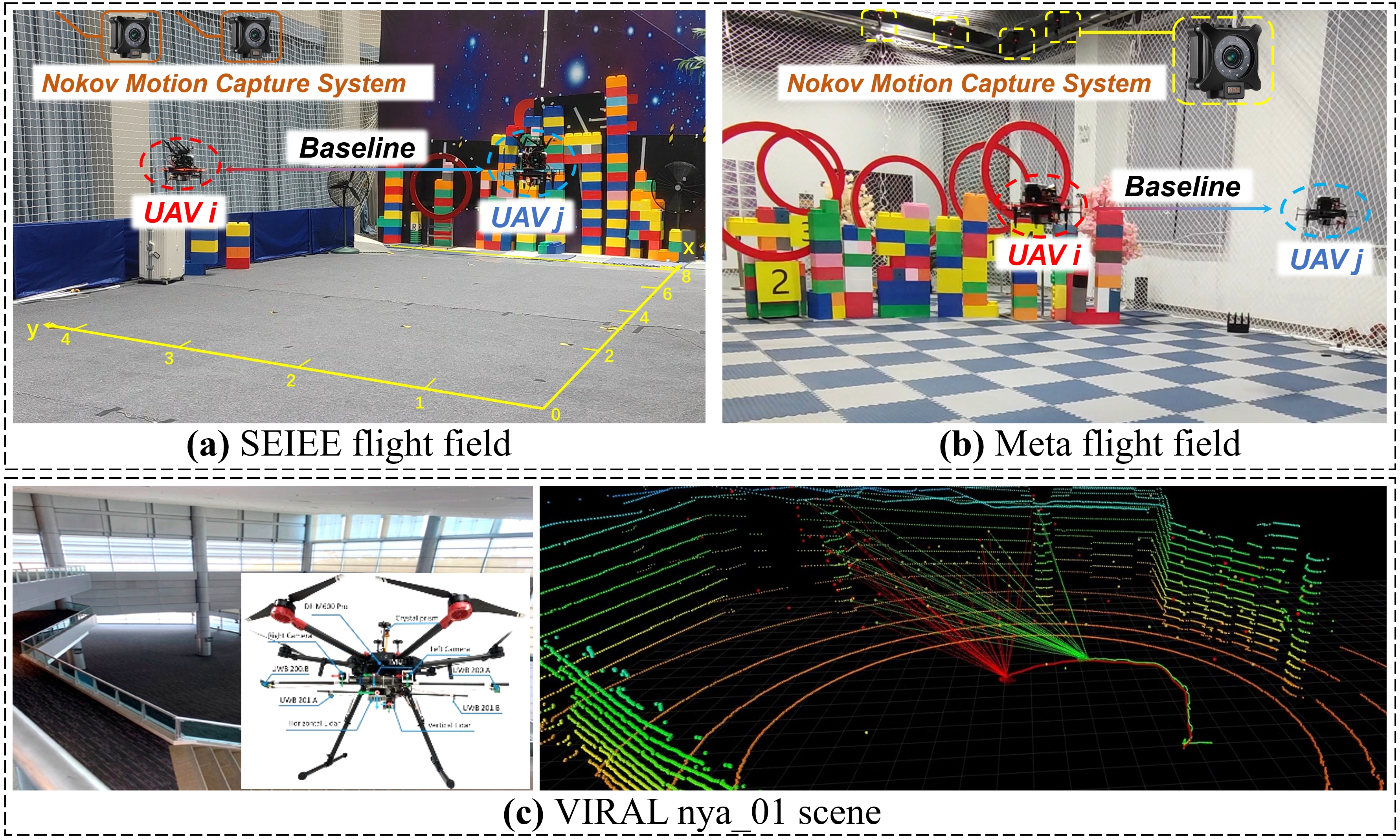}
  \caption{The flying experiments scenes for relative pose estimation. (a) shows the SEIEE flight field. (b) shows the Meta flight field. (c) shows nya\_01 scene in VIRAL dataset.}
\end{figure}

To evaluate the convergence performance under different baseline accuracy, we execute flight experiments with two UAVs in flight field, as shown in Fig.10. The two UAVs are autonomously controlled to execute rectangle flight trajectories. The baseline between the two UAVs is set to 3m along the Y-axis. The initial scene depth is 5m. The average flight speed is 0.5m/s. Both UAVs’ cameras are oriented forward. The coarse initialization of baseline is calculated by PnP. The 3D position of co-visible landmarks are provided by the master UAV $i$, then the slave UAV $j$ perform 3D-2D PnP to obtain the coarse relative pose with UAV $i$. Generally, the coarse initialization has position error within 0.8m. The initialization does not require the assistance of a motion capture system. However, to fully evaluate the convergence under different baseline errors, We manually add larger initial position error around ground truth by motion capture system from -2.0m to 2.0m with 0.5m step in X-Y-Z axises respectively. Since our Rel-MSCKF uses reprojection error to iteratively update the cloned states of UAV $j$, all the trajectories from various baseline initial errors can converge to the ground truth within 1.5s in Fig.10(c). The trajectories with larger initial errors require more convergence steps than those with smaller initial errors. Additionally, we perform another arc-shaped flight trajectory for both UAVs, as shown in Fig. 11. The baseline is set as 3m. Both UAVs’ cameras are oriented forward. The initial baseline error ranges from -2.0m to 2.0m in X-Y-Z axises. The average flight speed is 0.8m/s. The result shows all the trajectories can converge to the ground truth within 1s.

\begin{figure}[h]
  \centering
  \includegraphics[width=1.0\linewidth]{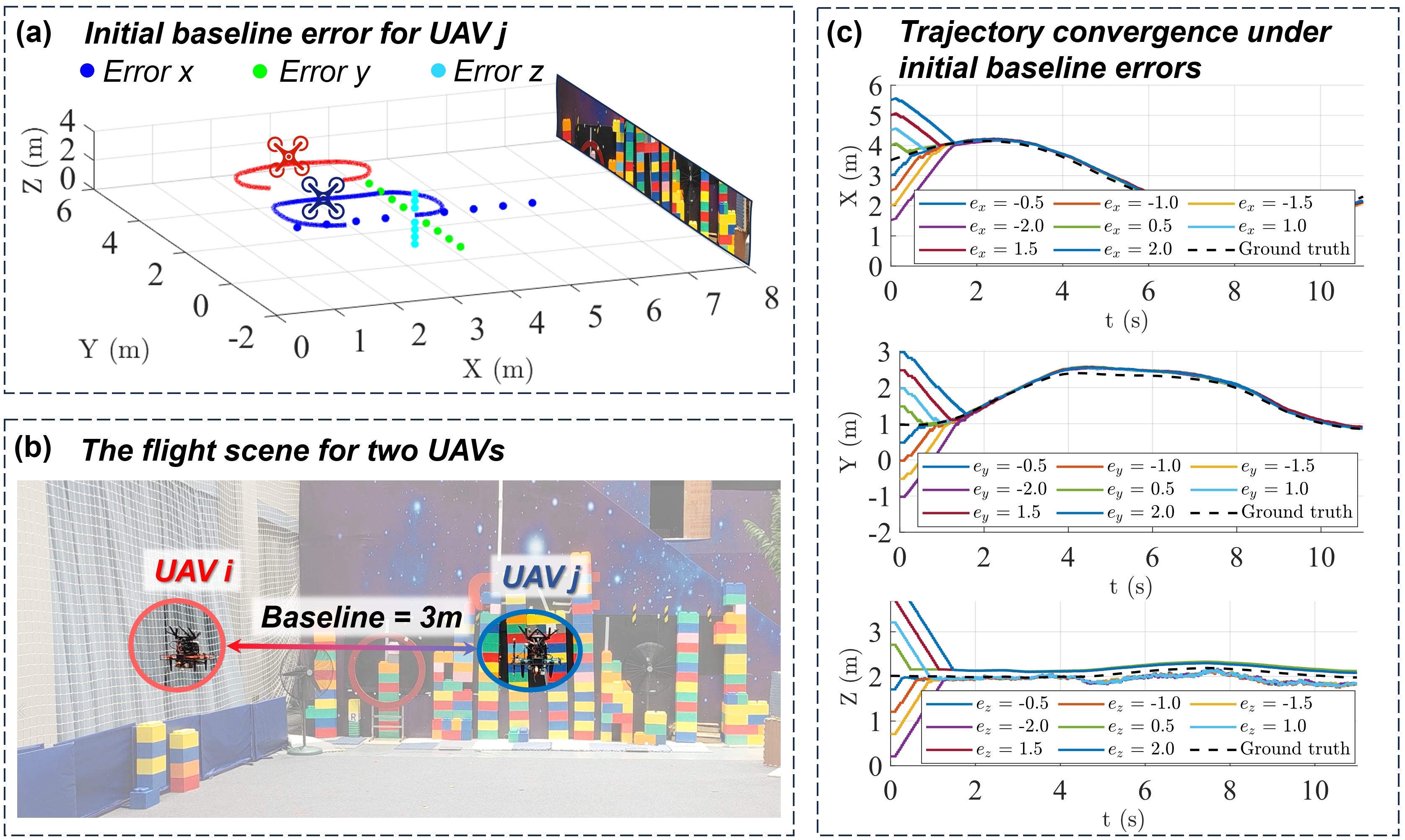}
  \caption{The convergence evaluation of Rel-MSCKF under initial baseline errors on flight field scene, where two UAVs follow rectangle trajectories.}
\end{figure}

\begin{figure}[h]
  \centering
  \includegraphics[width=1.0\linewidth]{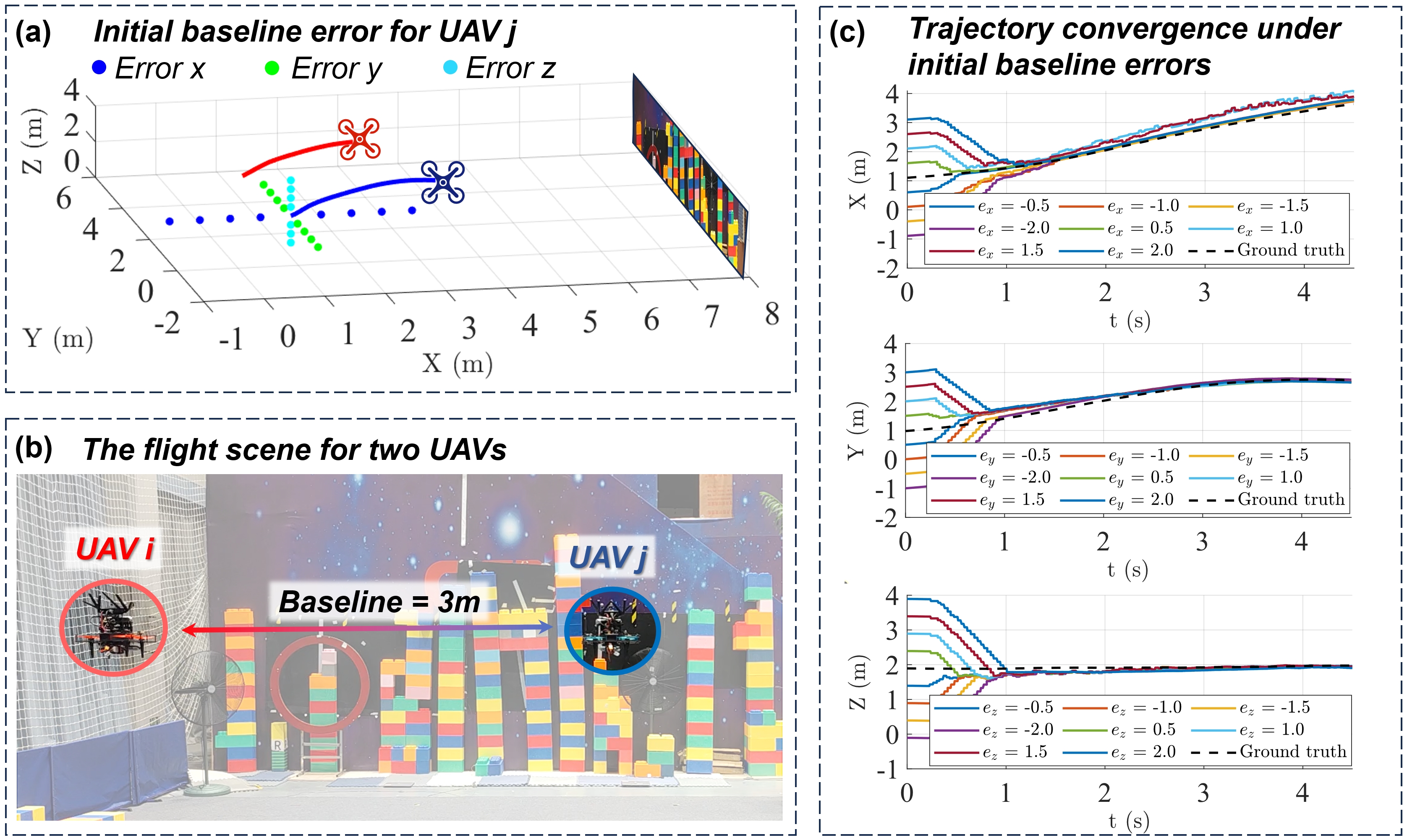}
  \caption{The convergence evaluation of Rel-MSCKF under initial baseline errors on flight field scene, where two UAVs follow arc trajectories.}
\end{figure}

To validate the correction performance of Rel-MSCKF under VIO drifting conditions, we present the trajectories of UAV $j$ from the above rectangle-shaped flight experiment in Fig.12. Compared to the ground truth trajectory, the VIO prediction has a smaller-scale estimation. Although the proposed Rel-MSCKF uses the drifted VIO for prediction, the update using common landmarks adjusts the pose closer to the ground truth. Due to airflow disturbances, the flight speed, as shown in Fig.12, is unstable and continuously fluctuates. Nevertheless, our Rel-MSCKF pose estimation achieves an RMSE of 0.133m upon convergence.

\begin{figure}[]
  \centering
  \includegraphics[width=1.0\linewidth]{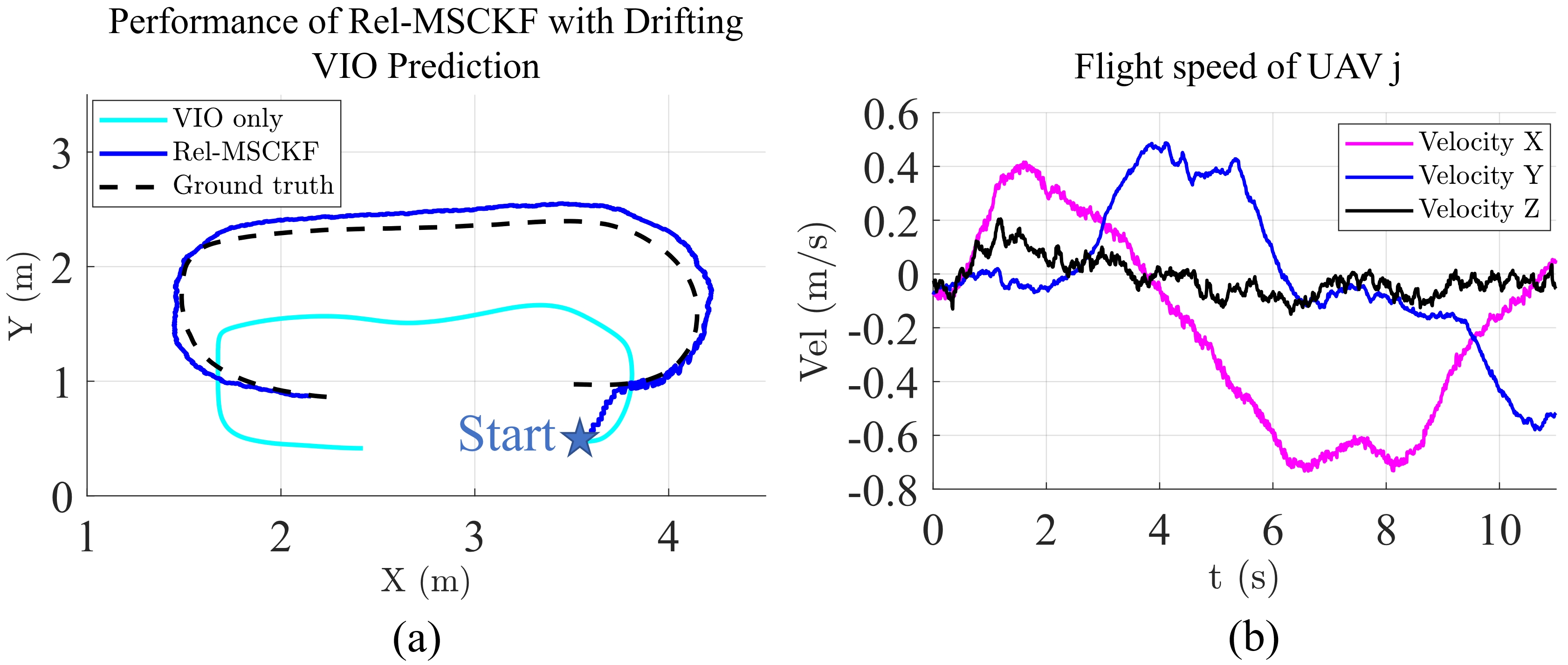}
  \caption{(a) shows The comparison of Rel-MSCKF trajectory, drifted VIO trajectory and ground truth. (b) plots the flight speed of UAV $j$ during the rectangle trajectory.}
\end{figure}

We compare the convergence and accuracy performance of our Rel-MSCKF with different feature association front-ends. The front-end modules are SSF, SS, SURF\_NR, and ORB\_NR respectively. The associated features from different front-end modules are equally processed by Rel-MSCKF. We adopt the above rectangle-shaped. Since the two UAVs execute the same rectangle trajectories, the plotted curves of relative trajectory are not obvious. We plot the estimated pose of UAV $j$ based on the relative pose against the ground truth of UAV $j$.

The estimated trajectories and ground truth are plotted in Fig.13. The error statistics are recorded in Table \textsc{III}. Compared to SS, SSF could fully utilize frames in the image stream, achieving a higher rate of feature observations and state updates within the same time duration. The high-rate pose estimation improves the convergence speed. The synchronization mechanism in SSF also guarantees accurate fusion with VIO prediction. For the estimation of SURF\_NR, the time-consuming feature association results in a low rate of feature observation, which leads to slower convergence. Due to the poor matching accuracy of ORB\_NR, the filtered valid feature matches are rare, leading to slow convergence.

The evaluations on the VIRAL dataset simulate two UAVs with a time offset of 3s on the single UAV dataset. The lidar data is used to generate depth images for UAV $i$. We add initial position noise to UAV $j$ and evaluate the estimated trajectories of UAV $j$ with four front-end methods. The estimated relative trajectories and ground truth are plotted in Fig.14. The N/E plot and error plot with covariance are shown in Fig.15. The trajectory with the SSF front-end converges more quickly than with the other front-ends. The pose error with the SSF front-end is also lower than with other front-end. Due to rapid changes in the field of view in this dataset, the two UAVs experience significant perspective differences. The insufficient matching accuracy of SURF and ORB results in false updates and drifts in the backend estimation. The error statistics are recorded in Table {\uppercase\expandafter{\romannumeral3}}. Similar to the results of self-constructed datasets, SSF with Rel-MSCKF achieves better estimation with the RMSE error of 0.113m than other front-ends. Detailed experiment scenarios can be found in our videos.

\begin{figure}[]
\centering
\includegraphics[width=1.0\linewidth]{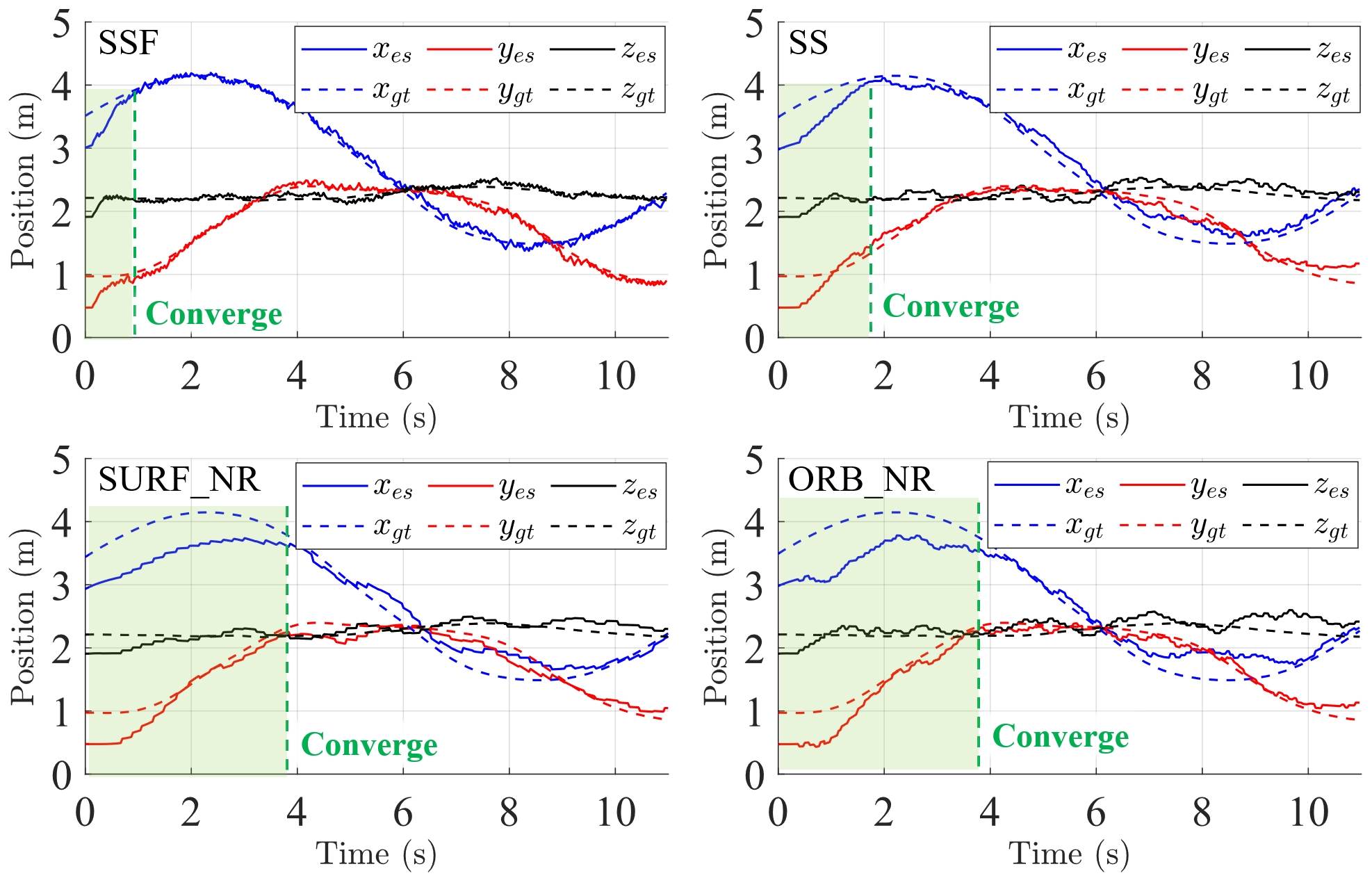}
\caption{The relative trajectory estimation by the front-end of SSF, SS, SURF\_NR, ORB\_NR with the Rel-MSCKF on SEIEE flight field.}
\end{figure}

\begin{figure}[]
  \centering
  \includegraphics[width=1.0\linewidth]{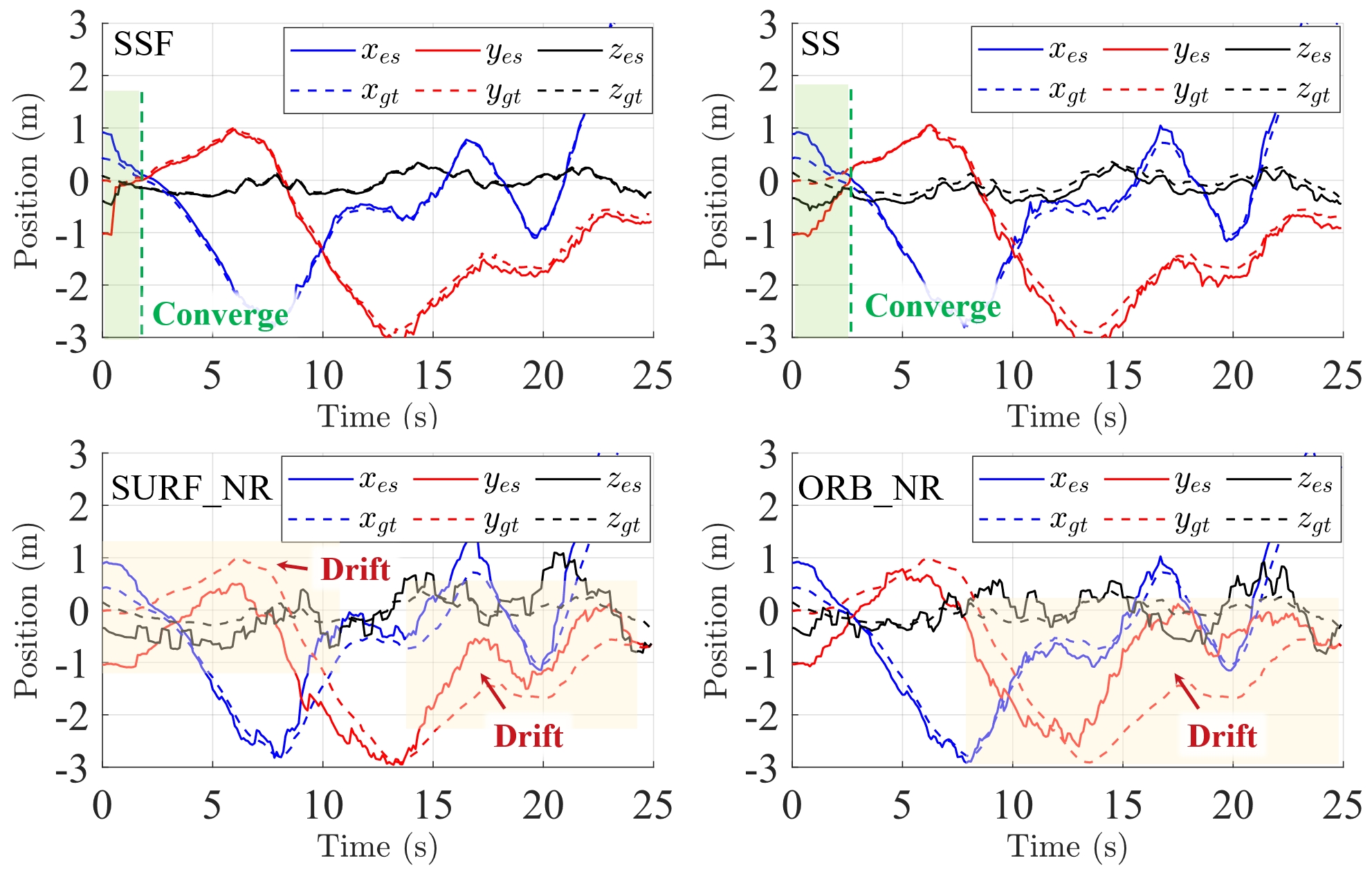}
  \caption{The experiment of simulated two UAVs with nya\_01 sequence in VIRAL dataset.}
\end{figure}

\begin{figure}[]
\centering
\includegraphics[width=1.0\linewidth]{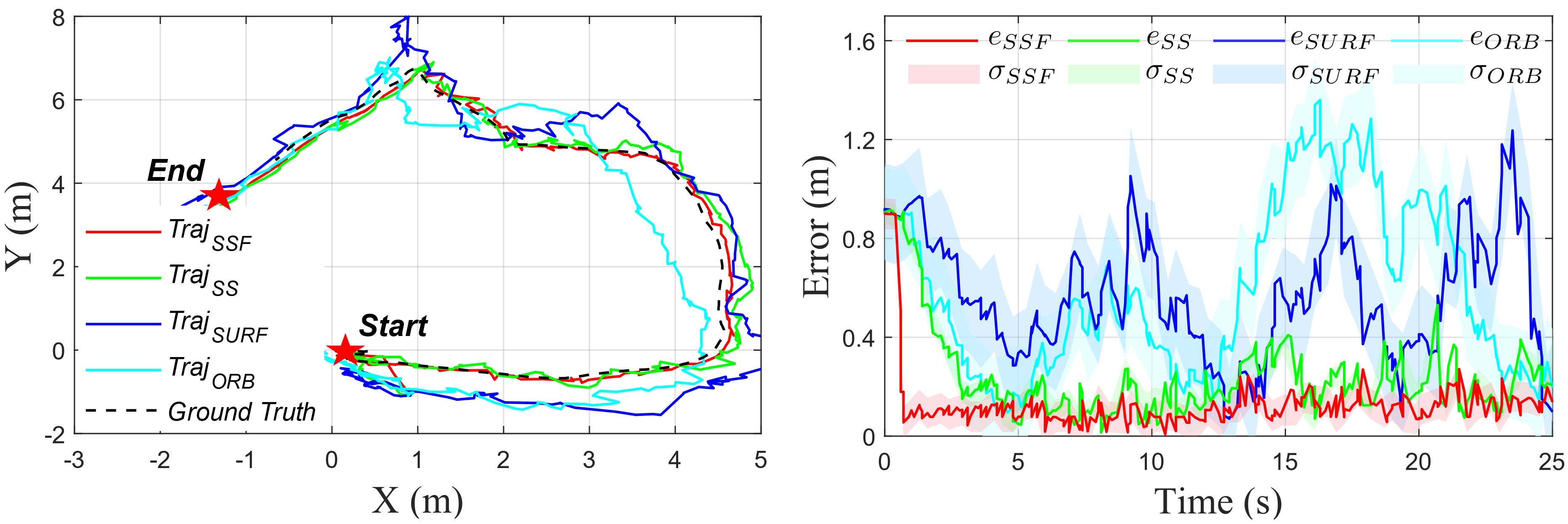}
\caption{The estimated trajectory of UAV j against ground truth and error curves with covariance on nya\_01 scene of VIRAL dataset.}
\end{figure}

\begin{table}[]
\centering
\caption{Pose Error Analysis on Flight Field and VIRAL Dataset}
\setlength{\tabcolsep}{2.8mm}{
\begin{tabular}{@{}llllll@{}}
\toprule
Dataset             & Front-end  & RMSE(m)         & STD(m)         & RMSE(\textdegree)        & STD(\textdegree)\\ \midrule
\multirow{4}{*}{\begin{tabular}[c]{@{}l@{}}Flight\\ Field\end{tabular}} & SSF        & \textbf{0.163}  & \textbf{0.112} & \textbf{1.387} & \textbf{0.188}  \\
                    & SS         & 0.298           & 0.156          & 1.707           & 0.192          \\
                    & SURF\_NR   & 0.404           & 0.196          & 2.265           & 0.823          \\
                    & ORB\_NR    & 0.444           & 0.214          & 2.520            & 0.577         \\ \midrule
\multirow{4}{*}{VIRAL} & SSF        & \textbf{0.113}  & 0.062 & \textbf{1.435} & \textbf{0.317}  \\
                    & SS         & 0.157           & \textbf{0.045}          & 1.513           & 0.458          \\
                    & SURF\_NR   & 0.351           & 0.198          & 2.856           & 1.553          \\
                    & ORB\_NR    & 0.932           & 0.188          & 3.103           & 1.561        \\ \bottomrule
\end{tabular}
}
\end{table}

We also compare the proposed Rel-MSCKF with the PGO-based batch optimization method. COVINS-G\cite{patel_covins-g_2023} is a typical PGO-based relative pose estimator. The comparison is performed on nya01 sequence of VIRAL public dataset. The trajectory snapshots of UAV $i$ and UAV $j$ are recorded in Fig.13. The position error between estimation and ground truth is plotted in Fig.14. Overall, the COVINS-G can align two trajectories after several PGOs. However, PGOs are only executed at 6s, 8s, 12s, and 17s, which has a low frequency of relative pose correction. COVINS-G normally collects a batch of keyframes that require sufficient translational variation and the abundant inlier feature matches ($>$100) with RANSAC before once PGO. The above requirements of frame selection enhance the robustness but decrease the frequency of PGO pose correction. Each PGO will cost about 330 ms ± 168 ms on the resource-constrained NVIDIA NX platform. The relative pose error between two UAVs cannot be corrected in time. In comparison, the proposed Rel-MSCKF could leverage reprojection residuals on each image frame to update the relative pose in time. The average run-time Rel-MSCKF is around 16 ms ± 8 ms. In our opinion, the PGO approaches are more suitable for globally consistent trajectory alignment. The heavy optimization computation is allowed to be executed in a non-real-time manner. Considering the background of collaborative stereo, we focus more on the current relative pose with real-time requirements. In addition, during continuous side-by-side flight, two UAVs will share continuously overlapping views. The LK-flow in the front-end has the advantage of continuously associating and predicting common features in real-time. Then, the high-rate common feature observation will support real-time state updates. Therefore, the proposed Rel-MSCKF is more appropriate for continuous and real-time relative pose estimation.

\begin{figure}[t]
  \centering
  \includegraphics[width=1.0\linewidth]{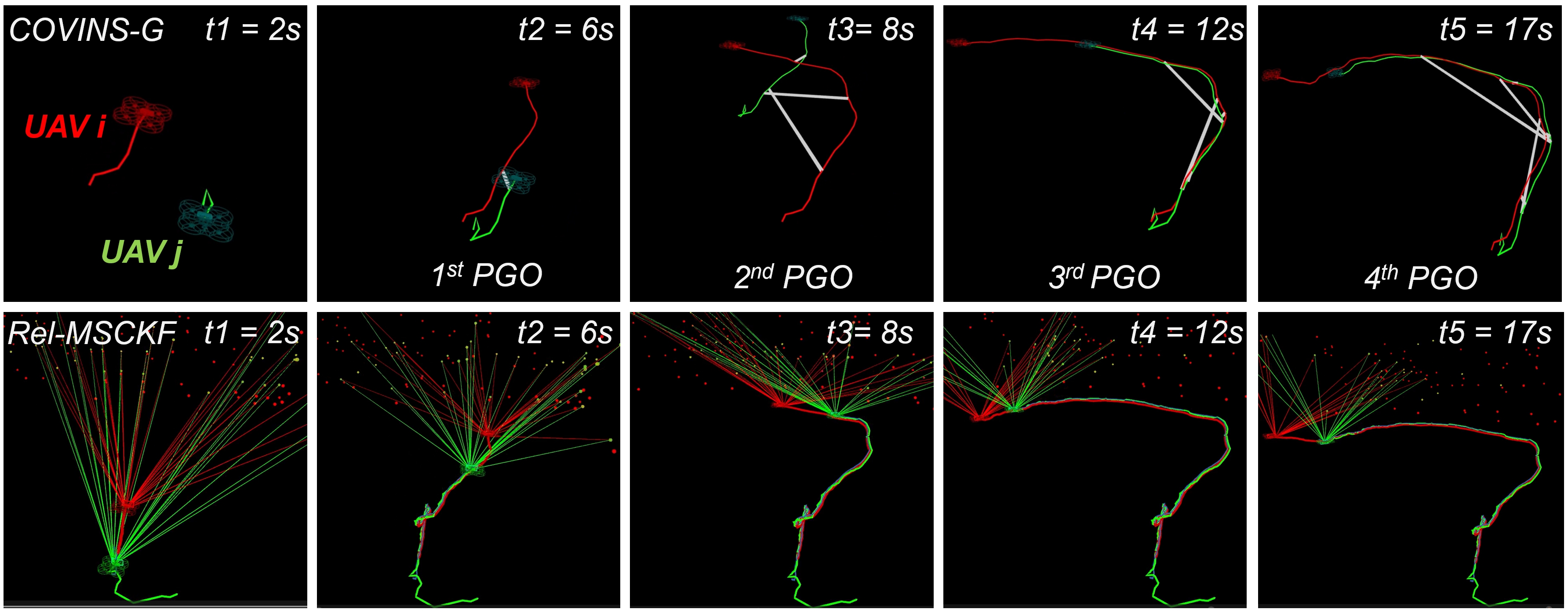}
  \caption{The trajectory snapshots of the Rel-MSCKF and COVINS-G.}
\end{figure}

\begin{figure}[]
  \centering
  \includegraphics[width=1.0\linewidth]{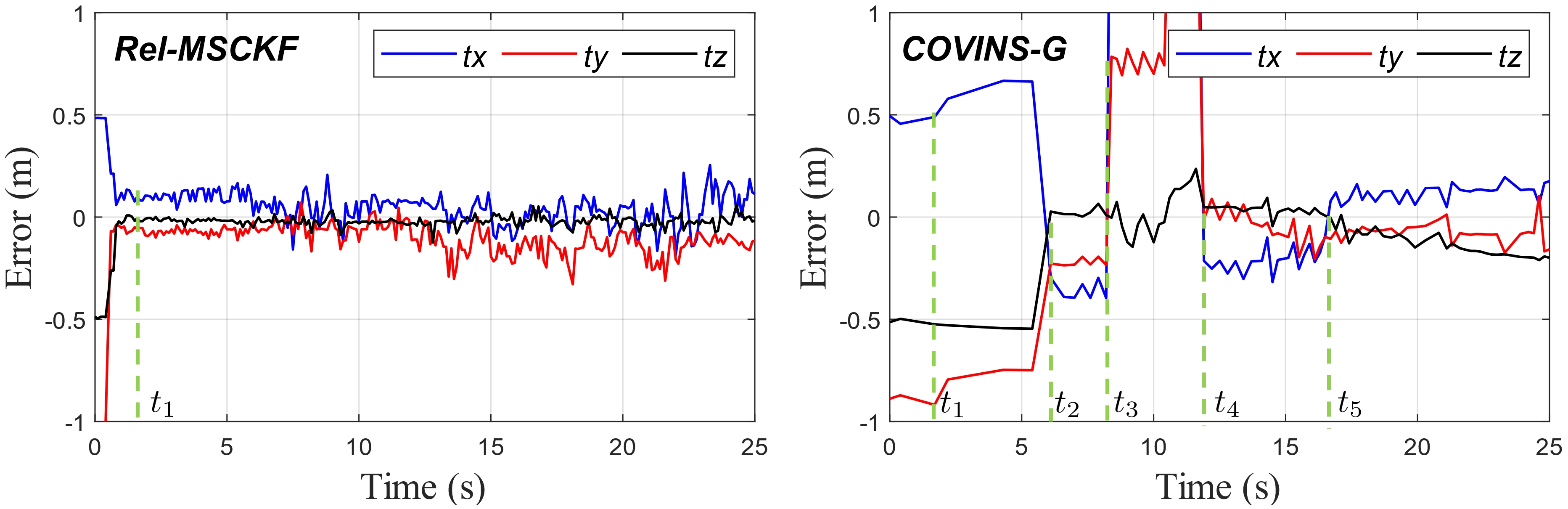}
  \caption{The trajectory error of the Rel-MSCKF and COVINS-G.}
\end{figure}

\subsection{Discussion on Spatial Configurations}
Different baselines, FOVs, and scene depths constitute various spatial configurations. In this part, we first perform geometric modeling of overlapping region considering baseline, FOV, and scene depth. Then, we assess the impact of the size of the overlapping region on the number of features extracted by the front-end. Subsequently, we carefully analyze the pose estimation results from flight experiments. Finally, we further investigate the state perturbation errors, considering both the number of features and different spatial configurations.

The geometry relationship between two UAVs is presented in Fig.18. The x-o-y coordinate system is established on the body of master UAV $i$. The scene depth $d$ is assumed to be parallel to the image plane of UAV $i$. The position of UAV $j$ is indicated by baseline $(b_x,b_y)$. Both UAVs are assumed to fly at the same height. Both cameras have an FOV of $\alpha$, with the yaw angle of UAV j's camera relative to UAV i being $\theta$. The view coverage $V_i$ of UAV $i$ at depth $d$ is from left endpoint $v_i^l$ to right endpoint $v_i^r$. Similarly, the view coverage $V_j$ for UAV $j$ is represented by $v_j^l$ and $v_j^r$. The overlapping region $V_{ij}$ is the intersection part of the two view coverages. We assume $v_j^l$ is left to $v_i^r$ temporarily and then extend for general condition. The overlapping area $V_{ij}$ can be calculated by
\begin{equation}
\begin{aligned}
    V_{ij} & = v_{i}^r - v_{j}^l \\
           & = d\cdot tan(\alpha/2) + (d-b_y)\cdot tan(\alpha/2 + \theta) - b_x
\end{aligned}
\end{equation}
The full view coverage $V_i$ of UAV $i$ at depth $d$ is $2d\cdot tan(\alpha/2)$. 
The proportion of the overlapping view $V_{ij}$ in full view coverage $V_i$ of UAV $i$ is represented by $Vp_i = V_{ij}/V_i$. In our flight field case, the FOV of camera $\alpha$ is 90${}^{\circ}$, depth $d$ is 8m. The proportion $Vp_i$ with different baselines and relative FOV orientation is plotted in Fig.19. The closer the depth, the smaller the overlapping proportion. Additionally, a smaller baseline will increase the overlapping proportion. 

\begin{figure}[]
  \centering
  \includegraphics[width=1.0\linewidth]{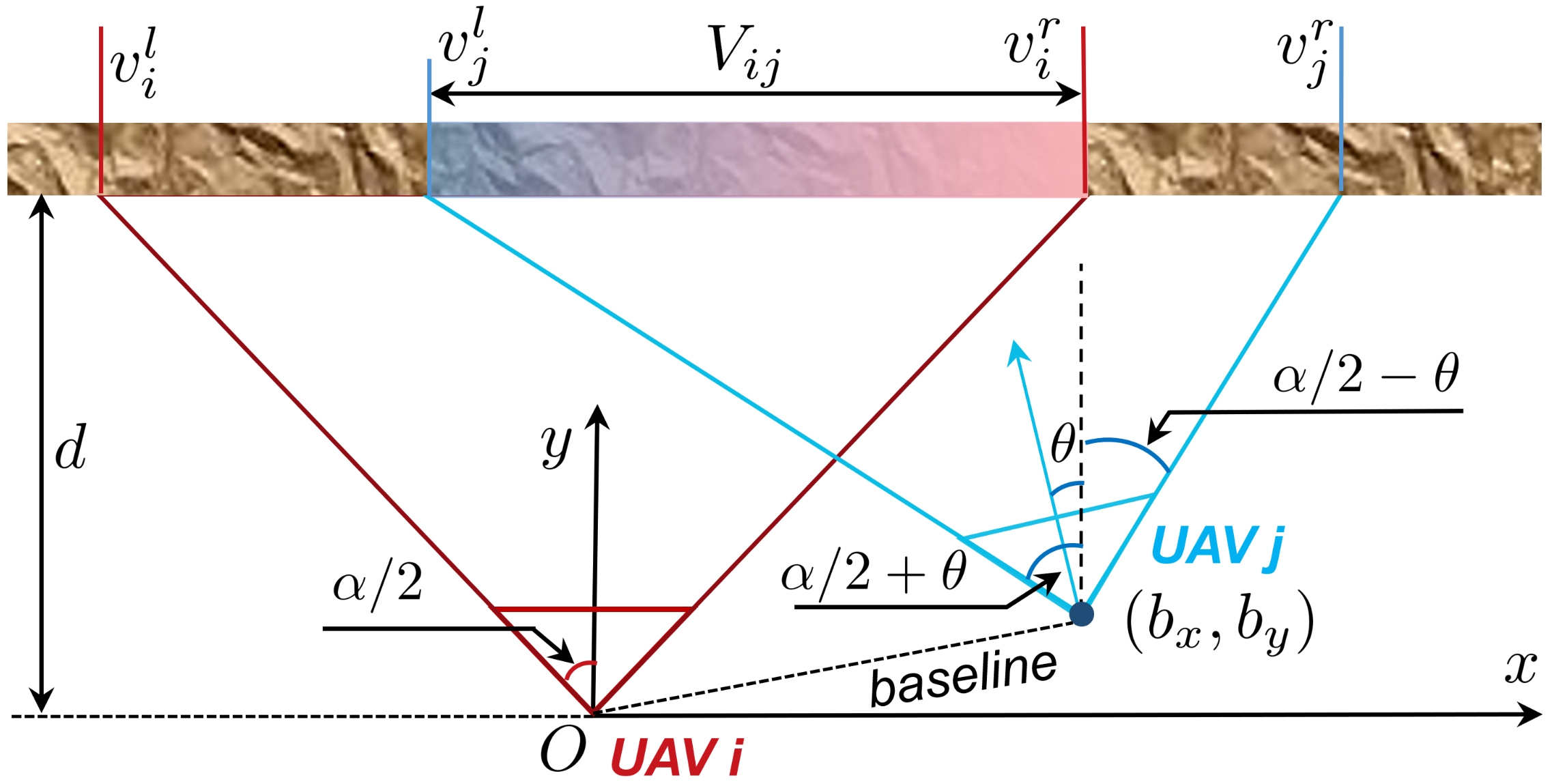}
  \caption{The spatial relationship between two UAVs, considering their baseline, field of view (FOV), and depth.}
\end{figure}

\begin{figure}[]
  \centering
  \includegraphics[width=1.0\linewidth]{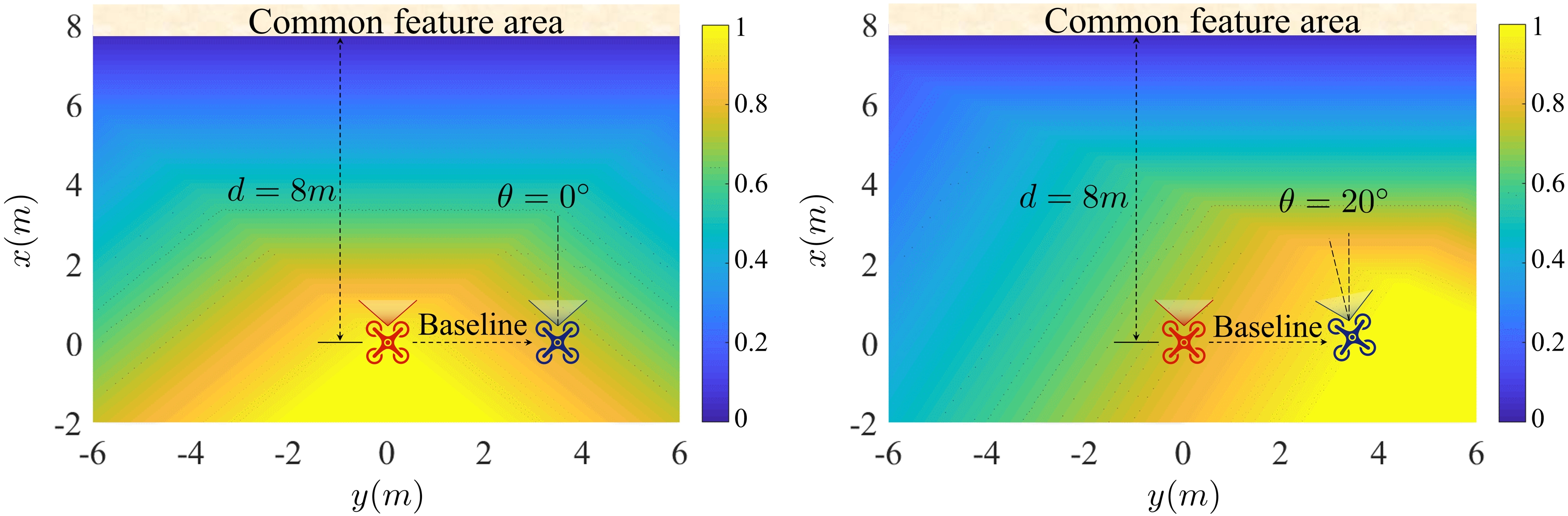}
  \caption{The proportion of the overlapping region in the full view coverage of UAV $i$. The overlapping proportion distributes between 0 and 1. The smaller baseline and larger scene depth improve overlapping proportion. The yaw angle $\theta$ adjustment of UAV $j$ also affects the overlapping proportion.}
\end{figure}

The proportion of the overlapping region within UAV i's view coverage affects the number of common features. 
To ensure a uniform distribution of features in the image, our front-end extracts features at evenly spaced grids on the image plane (40 px/grid in our case). Only the features that fall within the overlapping region in the full image can be considered common features. We conduct real-world flying experiments with different baselines, depths and yaw angles of slave UAV $j$. As shown in Fig.20, the experiments test scene depths ranging from 2m to 8m with baselines of 2m and 4m. The number of matches for yaw angle $\theta = -20{}^{\circ}, 0{}^{\circ}, 20{}^{\circ}$ are presented in Fig.21. Positive relative yaw angles increase the number of feature matches. Considering the above three factors, the scene diagram and the number of feature matches are shown in Fig.22. The smaller baseline (2m) increases overlapping proportion, leading to more feature matches. Generally, the number of feature matches is more sensitive to changes in baseline and yaw angle at smaller scene depths than at larger depths. In addition, small scene depth provide more feature matches than the large depth. The reason is that the height and width of the effective feature region are limited (a wall in our case). At large depths, the projection of the feature region relative to the full image will decrease, leading to fewer feature extractions.

In the real-world experimental setup, two aircraft synchronously perform a lateral flight of 1.5m to the left along the Y-axis in SEIEE flight field. The Y-axis trajectory ensures that the scene depth remains constant during flight. The results of relative pose estimation under varying baseline lengths ($B$), slave FOV angles ($\theta$), and scene depths ($D$) are presented in Table {\uppercase\expandafter{\romannumeral4}}. We can derive the following four analyses. First, the small depth will achieve more accurate pose estimation than the large depth. The small depth condition is highlighted in green ($D = 2m, B = 2m, \theta = 0{}^{\circ}, 20{}^{\circ}$). The large depth condition is highlighted in orange ($D = 8m, B = 2m, \theta = 0{}^{\circ}, 20{}^{\circ}$). On one hand, more feature points will engage the pose update at close depth; on the other hand, pixel matching noise induces smaller state errors at close distances. This will be further explained in the subsequent perturbation error analysis. Second, the condition with small baseline ($B=2m$) achieve better estimation than that with large baseline ($B=4m$). Third, the condition with FOV of $\theta = 20{}^{\circ}$ at the large baseline of 4m generally has more accurate estimation than $\theta = 0, -20{}^{\circ}$ due to the larger overlapping proportion. However, with the smaller baseline of 2m, the condition with FOV of $\theta = 0{}^{\circ}$ performs better than the other FOV angles. Fourth, the data highlighted in red with $-$ ($D = 2m, B = 4m, \theta = -20{}^{\circ}, 0{}^{\circ}$), due to divergent relative FOV and long baseline, lack sufficient overlapping fields of view. As a result, pose estimation is not performed.

\begin{figure}[]
  \centering
  \includegraphics[width=1.0\linewidth]{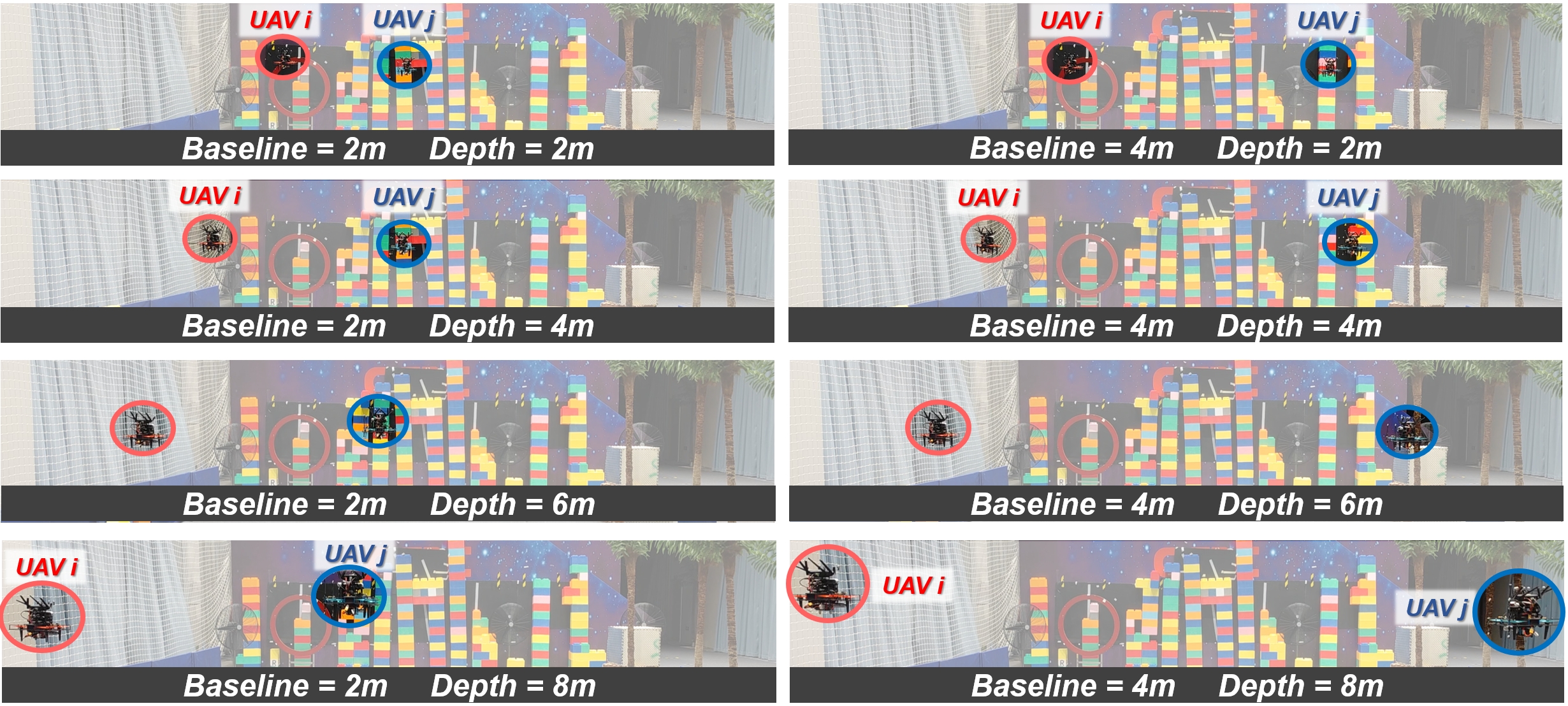}
  \caption{The experiments with different baselines and scene depths.}
\end{figure}

\begin{figure}[]
  \centering
  \includegraphics[width=1.0\linewidth]{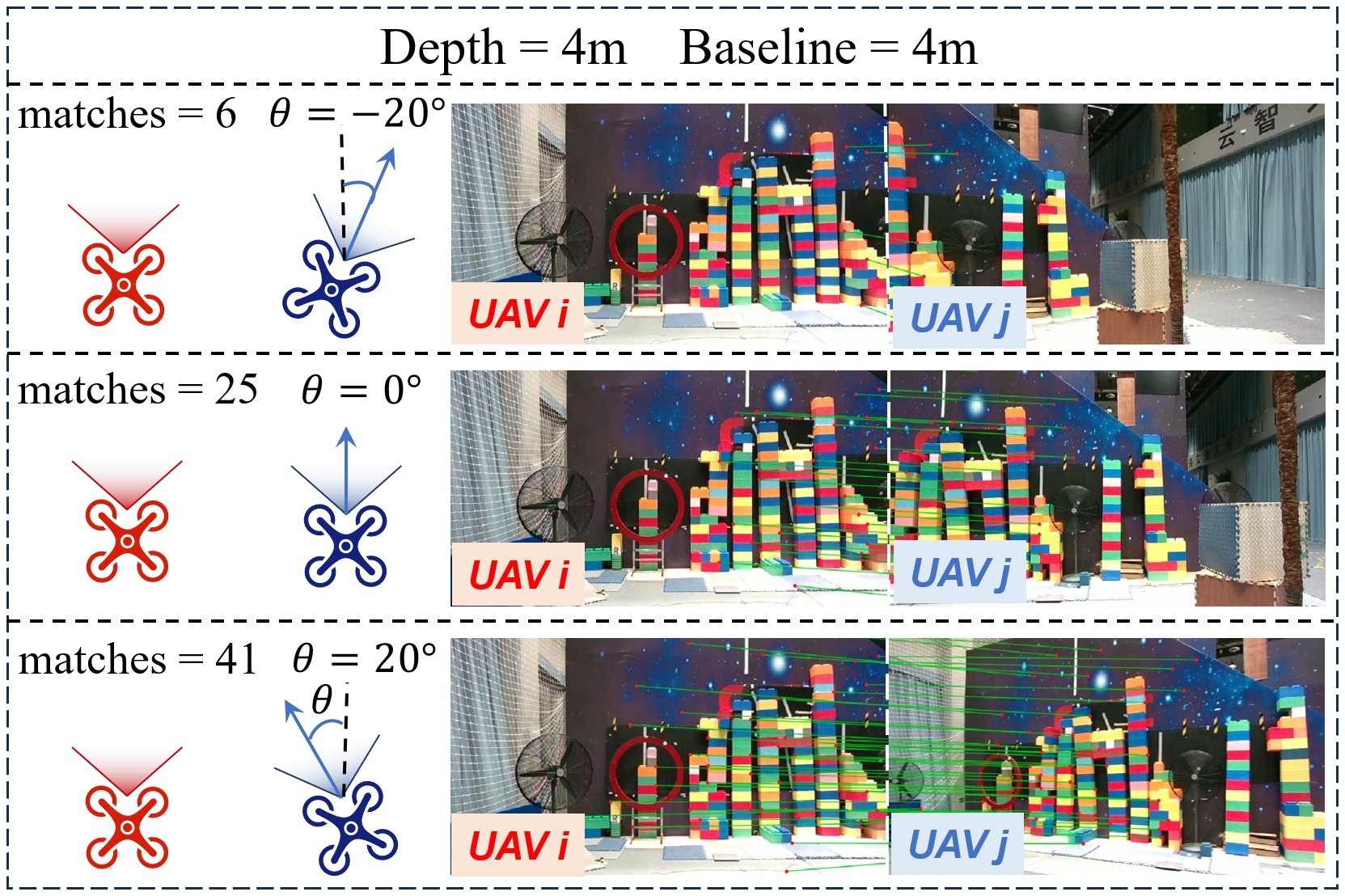}
  \caption{The number of matches under different yaw angles of UAV $j$.}
\end{figure}

\begin{figure}[]
  \centering
  \includegraphics[width=1.0\linewidth]{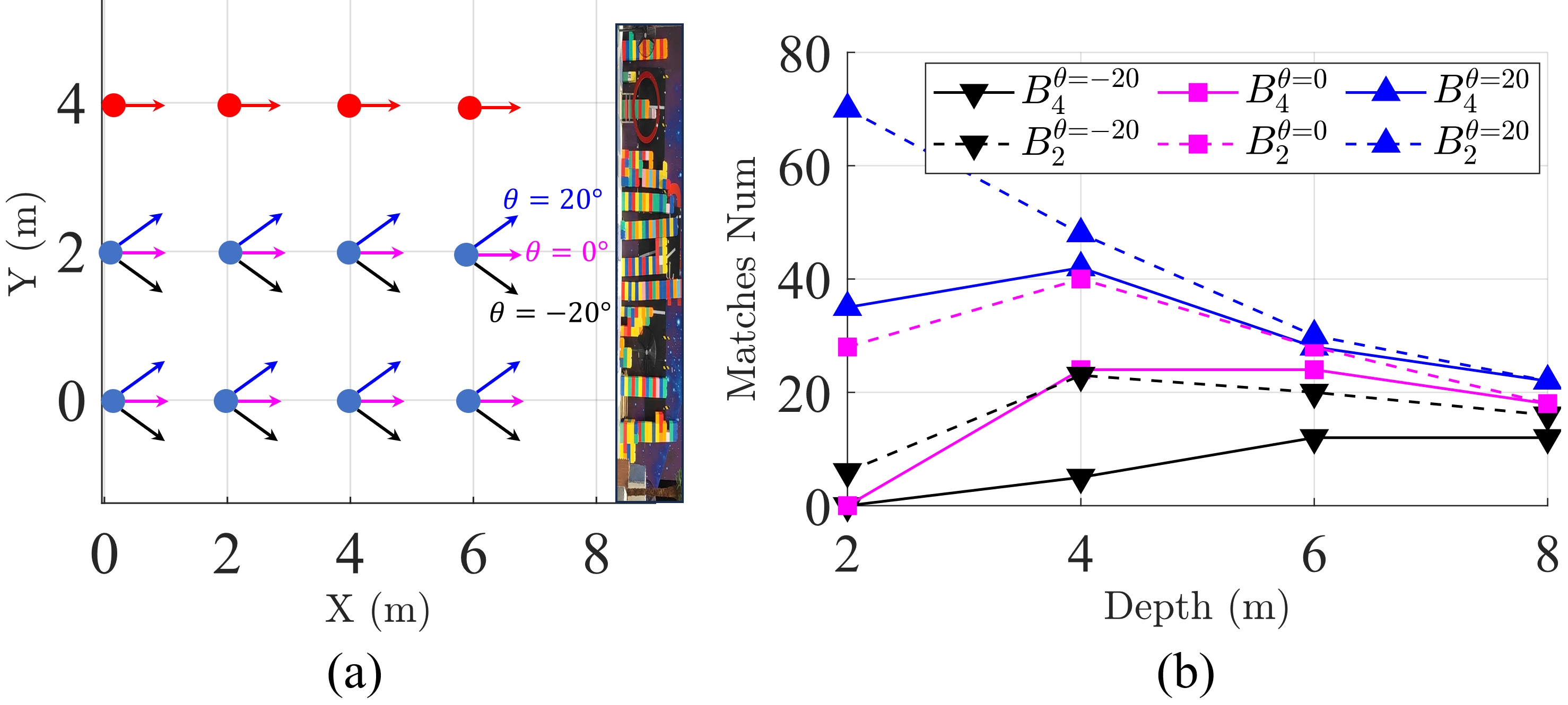}
  \caption{The number of matches under different baselines, scene depths, and yaw angles. The yaw angle of $\theta = -20{}^{\circ}, 0{}^{\circ}, 20{}^{\circ}$ are colored by black, pink, and blue respectively. The baseline of 2m is presented with dashed line and 4m for solid line.}
\end{figure}

\begin{table}[]
  \centering
  \caption{Pose Error Analysis under Different Baseline Lengths, FOV Angles, and Scene Depths}
  \setlength{\tabcolsep}{1.0mm}{
    \begin{tabular}{cccccccc}
      \toprule
      \multicolumn{2}{l}{}             & \multicolumn{3}{c}{Position RMSE(m)}                & \multicolumn{3}{c}{Orientation RMSE(\textdegree)}  \\ \cline{3-5} \cline{6-8}
      D(m)           & B(m) & $\theta=-20{}^{\circ}$ & $\theta=0{}^{\circ}$ & $\theta=20{}^{\circ}$ & $\theta=-20{}^{\circ}$ & $\theta=0{}^{\circ}$ & $\theta=20{}^{\circ}$ \\  \hline
      \multirow{2}{*}{2} & 2           & 0.231        & \textcolor{deepgreen}{0.118}      & \textcolor{deepgreen}{0.113}       & 2.331           & 1.301         & 1.285          \\
                         & 4           & \textcolor{red}{-}      & \textcolor{red}{-}        & 0.134       & -        & -           & 1.522  \\ \hline
      \multirow{2}{*}{4} & 2           & 0.135        & 0.124      & 0.125       & 1.536           & 1.361         & 1.401      \\
                         & 4           & 0.189        & 0.142      & 0.133       & 1.884           & 1.665         & 1.457      \\ \hline
      \multirow{2}{*}{6} & 2           & 0.161        & 0.149      & 0.146       & 1.721           & 1.682         & 1.425       \\
                         & 4           & 0.197        & 0.152      & 0.162       & 1.957           & 1.634         & 1.756       \\ \hline
      \multirow{2}{*}{8} & 2           & 0.178        & \textcolor{orange}{0.177}      & \textcolor{orange}{0.183}       & 1.768           & 1.752         & 1.865          \\
                         & 4           & 0.198        & 0.195      & 0.192       & 2.105           & 1.988         & 1.934          \\ \bottomrule
      \end{tabular}
  }
\end{table}

Finally, we perform the error perturbation analysis of pose estimation considering the feature observations and spatial configurations. Since the pose is updated through feature reprojection, we investigate how reprojection error affects the accuracy of pose estimation. The pixel error $(\Delta u,\Delta v)$ in a single feature measurement translates to a position perturbation error $(\Delta X, \Delta Y, \Delta Z)$ of the camera, as described by the following projection Jacobian:

\begin{equation}
\left[\begin{array}{c}
  \Delta u \\
  \Delta v
  \end{array}\right]=\left[\begin{array}{ccc}
  \frac{1}{Z} & 0 & -\frac{X}{Z^2} \\
  0 & \frac{1}{Z} & -\frac{Y}{Z^2}
  \end{array}\right]\left[\begin{array}{l}
  \Delta X \\
  \Delta Y \\
  \Delta Z
  \end{array}\right]
\end{equation}

We stack the pixel errors of $n$ features and construct an overdetermined system of equations $\boldsymbol{A}_{2n \times 3} \boldsymbol{x} = \boldsymbol{b}_{2n}$ to estimate the position perturbation errors. $\boldsymbol{x}$ is $(\Delta X, \Delta Y, \Delta Z)$ and $\boldsymbol{b}_{2n}$ is the stacked pixel errors. We solve this least squares problem using Singular Value Decomposition (SVD). Since the projection Jacobian is influenced by feature positions in camera of UAV $j$, the scene depth and baseline directly affect the projection Jacobian. The number of features directly affects the data scale for the least squares estimation. In our experiment, we generate Gaussian pixel errors with a mean of zero and a standard deviation of 5 pixels. The number of cross-camera feature matches and the position perturbation errors are shown in Fig.23. As the depth increases, the projection Jacobian decreases, causing the same pixel noise to result in greater position perturbations. Additionally, an increase in the number of feature matches reduces the impact of each pixel error on the camera position perturbation.

\begin{figure}[]
  \centering
  \includegraphics[width=1.0\linewidth]{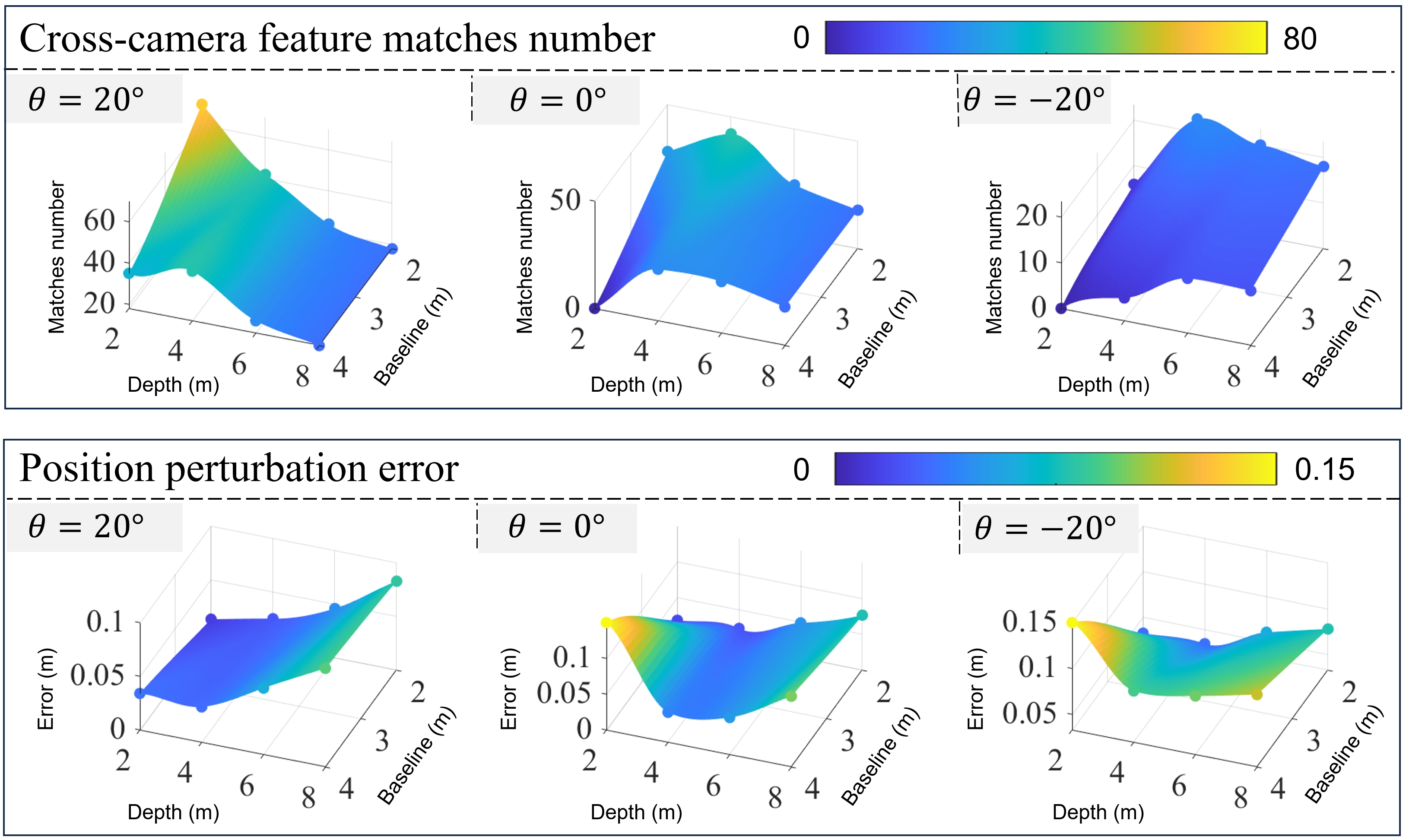}
  \caption{The number of cross-camera feature matches and according position perturbation under different baselines, depths, and yaw angles. The perturbation error is norm of $(\Delta X, \Delta Y, \Delta Z)$. For spatial locations with zero feature matches, we assign the maximum error of 0.15m for visualization in the plot.}
\end{figure}

\subsection{Robustness Evaluation of Collaborative UAV Systems}
The proposed collaborative UAV system may encounter challenges with asynchronous image acquisition, communication dropouts, and visual occlusions. In this section, we evaluate the robustness of the collaborative UAV system.

We first introduce the implementations to tackle the challenge of asynchronous image acquisition for two UAVs. Precisely synchronizing image exposure relies on a hardware trigger for both cameras, which is currently challenging for two independent cameras. We have made every effort to minimize the impact of asynchrony on state estimation. First, We carefully select independent WiFi Mesh network (TP-Link WDR7650), which provides reliable high-bandwidth, low-latency communication within a 15-meter range in open, unobstructed conditions. Second, we synchronize the system clocks of the two onboard computers using Network Time Protocol (NTP) to achieve a precision of approximately 1 ms. Third, we employ a queue buffer to pair images with the closest timestamps from both UAVs. Fourth, we currently focus on static environments. For the asynchronous images captured by the two cameras, we first compute the asynchronous relative poses, and then utilize VIO incremental propagation to obtain the relative pose at the same timestamp. This idea is inspired by a study \cite{multiSensorFusion2013}.

The detailed process is shown in Fig.24. Normally, the image from the master UAV is paired with the closest timestamp from the slave UAV. With a frame interval of 0.033 seconds, we intentionally pair the image with one captured 5 frames earlier, corresponding to a 0.16-second delay in the slave UAV's image. We first update the asynchronous relative pose of the slave UAV at 0.17s and the master UAV at 0.33s. Since the VIO prediction of slave UAV has already reached 0.33s, we can accumulate the VIO increments from 0.17s to 0.33s to fast propagate the relative pose at 0.33s. Thus, we can estimate the relative pose simultaneously despite asynchronous image acquisition.

We further evaluate the impact of different time asynchronization on estimation performance. As shown in Fig.25, we test the time difference of the asynchronous images with 0.00s, 0.06s, 0.10s, 0.16s, 0.22s, and 0.33s. We select the rectangle trajectory in SEIEE flight field with an initial baseline error.
For cases with significant asynchrony, such as 0.33s, where the slave UAV's timestamp is older, the total fast propagation of VIO values becomes larger. During the convergence process from 0s to 1s, the fast propagation is based on an inaccurate relative pose, leading to more errors due to larger propagation. However, once convergence is achieved (1s $\sim$ 2.5s), the fast propagation is based on accurate relative pose, and the impact of different asynchrony levels becomes negligible. Experiments show the proposed system can resist asynchronization of image acquisition.

\begin{figure}[]
  \centering
  \includegraphics[width=1.0\linewidth]{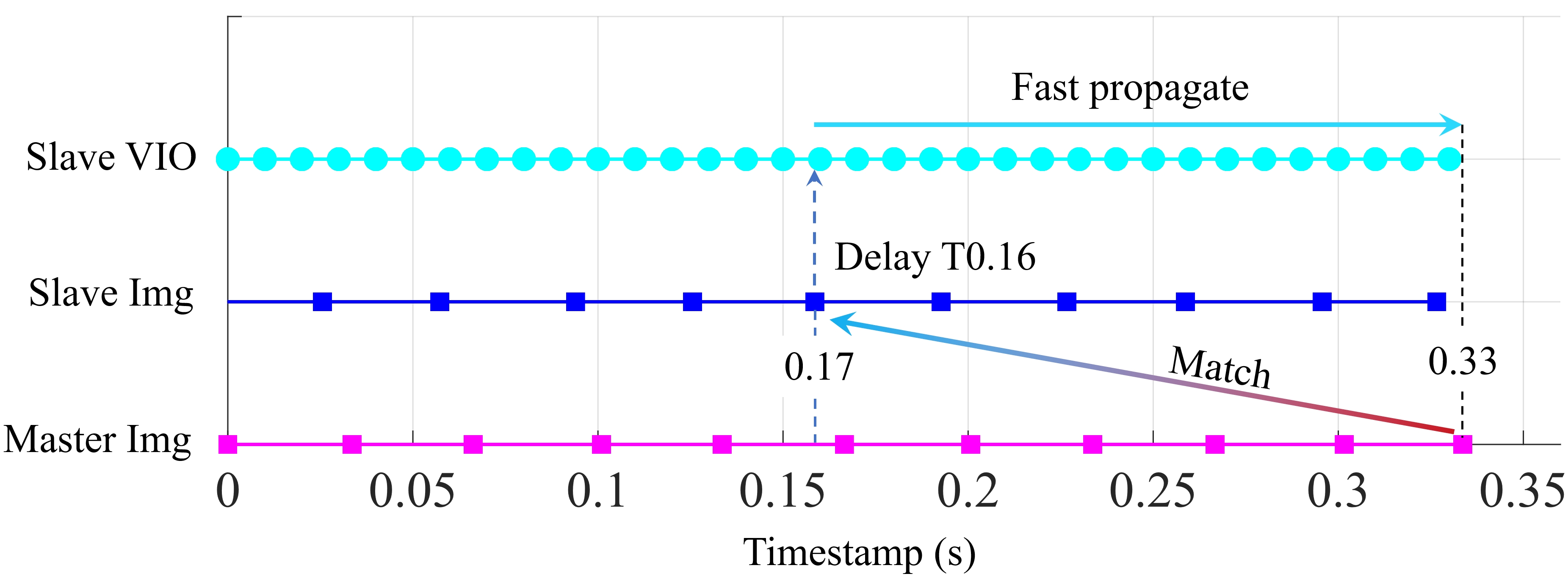}
  \caption{The illustration for asynchronous image match and fast propagation. The image timestamps of master UAV $i$ and slave UAV $j$ are plotted by pink and blue squares. The VIO timestamps of slave are cyan circle.}
\end{figure}

\begin{figure}[]
  \centering
  \includegraphics[width=1.0\linewidth]{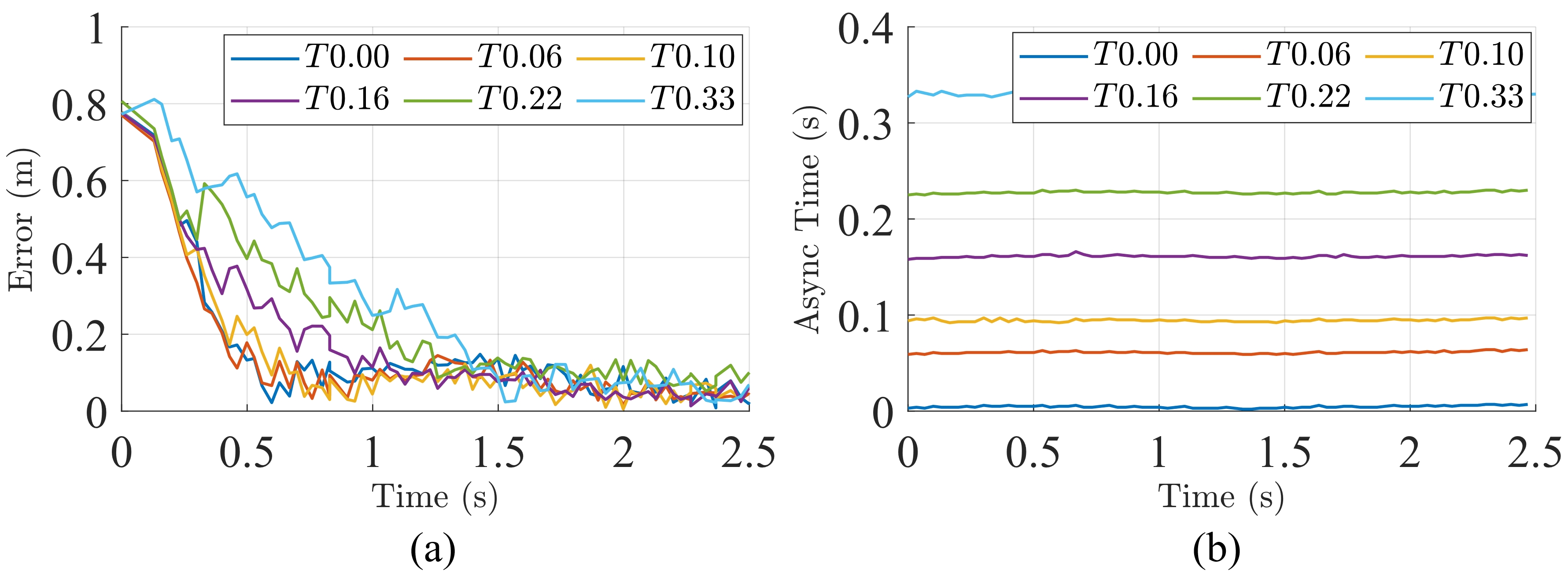}
  \caption{(a) shows the trajectory error convergence under different time asynchrony from 0s to 2.5s. (b) records the time difference of the asynchronous images between two cameras.}
\end{figure}

Second, we discuss the communication between two UAVs. The collaborative stereo normally relies on a well-established network for data exchange. In Fig.26, we have explored a feasible and stable hardware approach for communication. Each onboard computer is connected to a mesh router by a Gigabit Ethernet cable. The data from the onboard computer can be exchanged between meshed routers with high bandwidth and low latency. For quantitative evaluation, the image stream is 640 $\times$ 480 at 30 Hz and the SuperPoint is extracted with 150 keypoints at 13 Hz. Thus, the total bandwidth of exchanged data is around 2.1 MB/s, including the compressed gray image 50 KB $\times$ 30 Hz, 150 key points with the feature ID, pixel location and SuperPoint descriptor 150 $\times$ (1+8+256) B $\times$ 13 Hz, the local odometry pose 28 B $\times$ 30 Hz. We also validate the robustness under communication dropouts. The experiment is performed in the Meta flight field. UAV $i$ hovers, and UAV $j$ autonomously flies an 8-shaped trajectory. Both camera views of the UAVs remain forward-facing. The manually added communication dropouts will temporarily interrupt relative pose estimation. But once the communication recovers, the relative pose restarts to predict with historical increments of VIO and update with common features.

\begin{figure}[]
  \centering
  \includegraphics[width=1.0\linewidth]{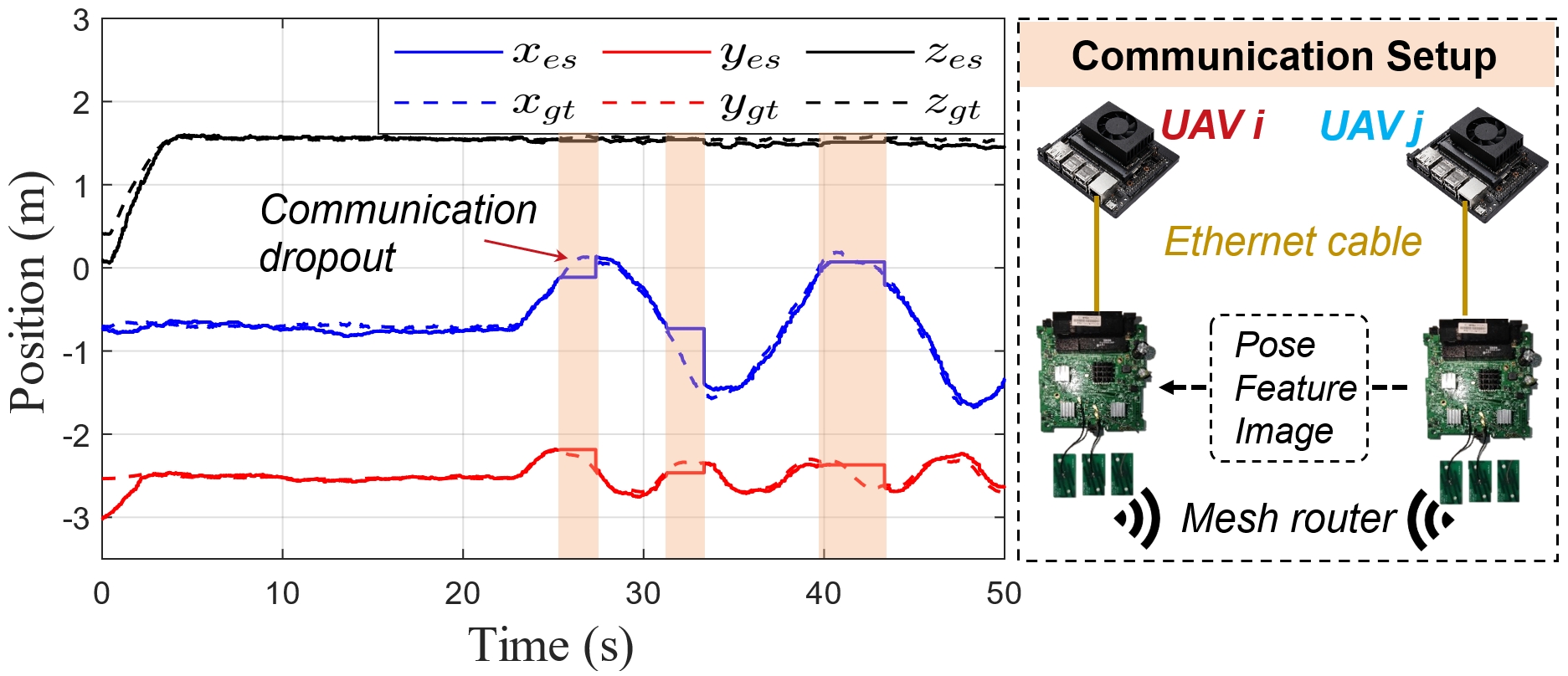}
  \caption{The condition of temporary communication dropouts and the communication setup.}
\end{figure}

\begin{figure}[h]
  \centering
  \includegraphics[width=1.0\linewidth]{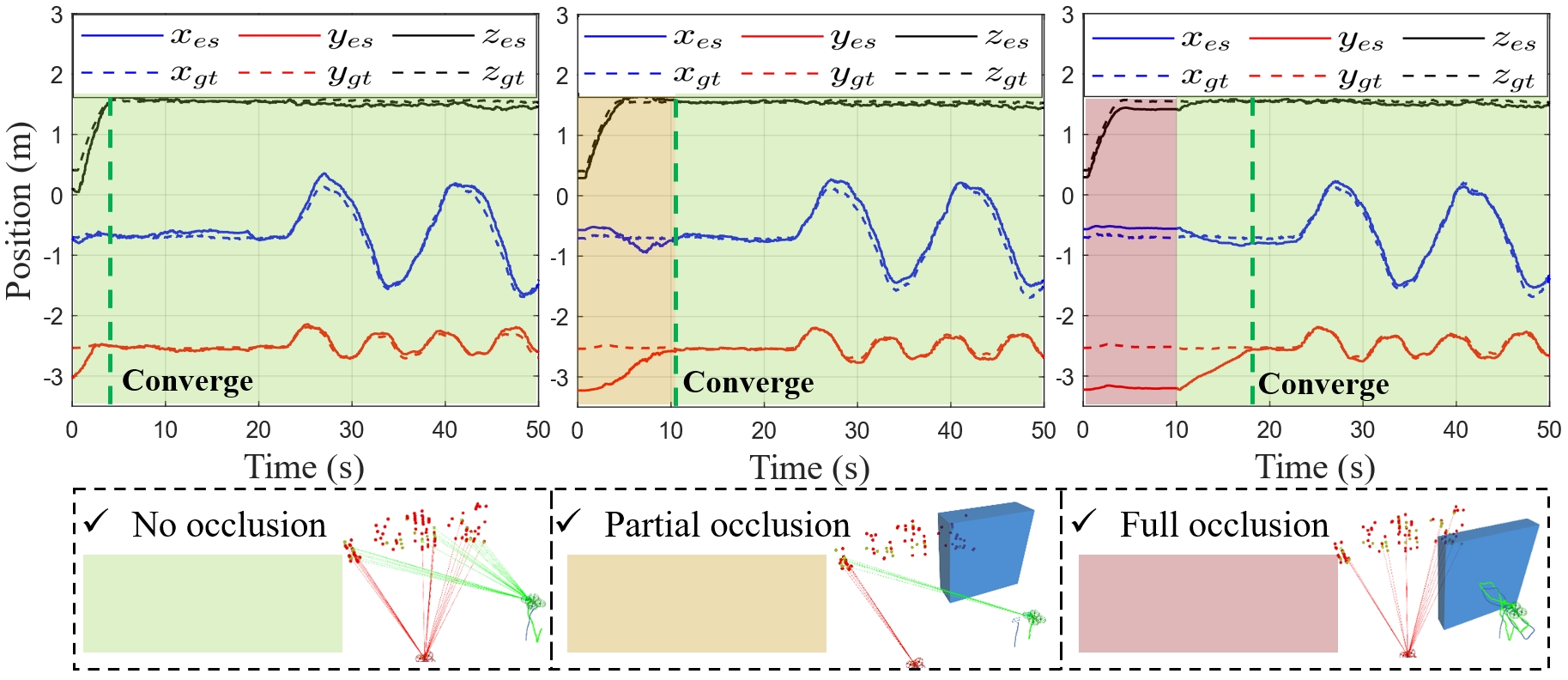}
  \caption{The estimated relative trajectories under visual occlusions.}
\end{figure}

Third, we execute experiments considering visual occlusions of common features. Three conditions, including no occlusion, partial occlusion and full occlusion, are presented for comparison. The trajectories with different visual occlusions are plotted in Fig.27. If there is no visual occlusion, sufficient common features will continuously impose enough reprojection error to correct the relative pose. Thus, the estimated trajectory can converge to ground truth quickly. If there is partial visual occlusion (0 $\sim$ 10s), the rare common features are intermittently observed. Thus, the rare and discontinuous feature reprojections will slow down the update progress of Rel-MSCKF. The speed of trajectory convergence is slower than that with no occlusion. If there is a full occlusion of common features (0 $\sim$ 10s), two UAVs only predict their poses in local coordinates without information about the neighbors. Thus, their relative pose cannot be corrected to the ground truth during the period of full occlusion. However, if the full occlusion happens in the middle of the trajectory, two UAVs can still predict relative pose based on the last relative pose and individual VIO propagations through communication. The yaw and position of slave UAV j could be controlled to recover the overlapping view with UAV i. Once the common features are available, the relative pose restarts to be corrected by Rel-MSCKF.

\begin{figure}[h]
  \centering
  \includegraphics[width=1.0\linewidth]{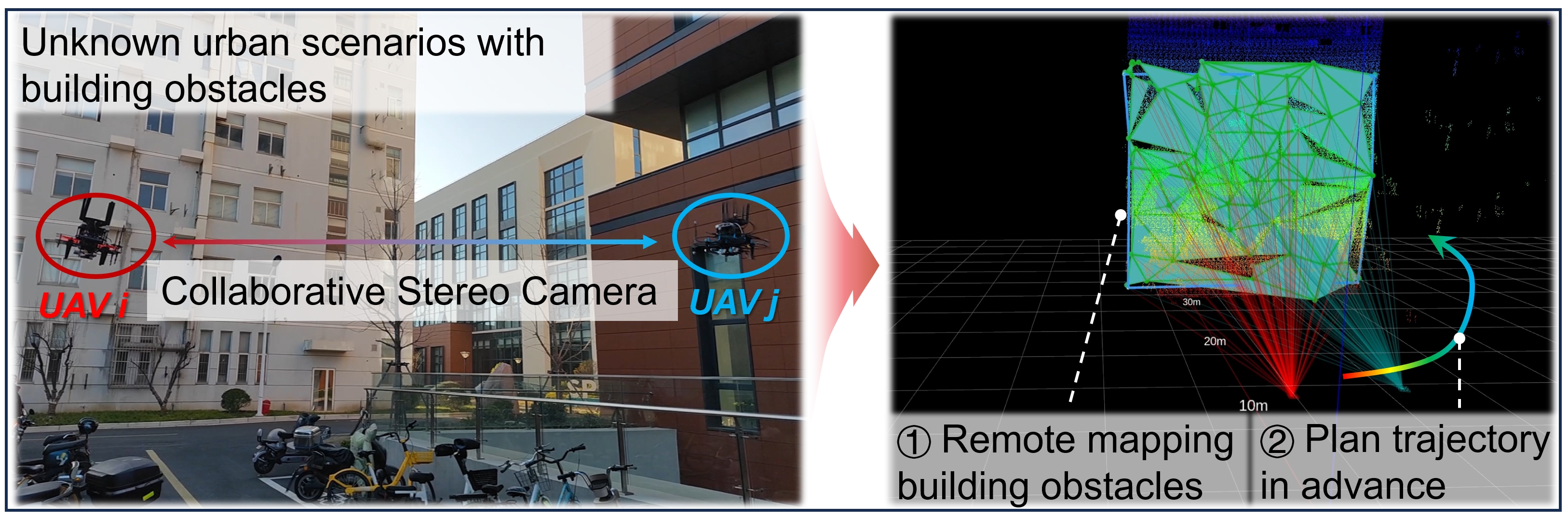}
  \caption{The future work of collaborative stereo for remote mapping and navigation in urban scenarios.}
\end{figure}

\section{Conclusion and Future Work}
In this paper, we first analyze the real-world challenges in aerial collaborative stereo during dynamic flight, which are the highly real-time requirement for the cross-camera feature association and the relative pose estimation of two UAVs. To address the first challenge, we propose a dual-channel algorithm to improve real-time performance on cross-camera feature association in large view disparities. Compared with existing single-channel feature association approaches, the dual-channel algorithm can reach 30 Hz on NVIDIA NX platform. For the second challenge, A lightweight Rel-MSCKF is developed to estimate relative pose, which outperforms the PGO-based approaches in real-time performance. 
The real-world flight experiments validate the effectiveness of the dual-channel feature association and Rel-MSCKF algorithms. Feature association and pose estimation are comprehensively analyzed across various spatial configurations. The robustness of the proposed system is further evaluated under challenges such as asynchronous image acquisition, communication dropouts, and visual occlusions.

The two foundational modules addressed in this paper pave for high-level tightly collaborative perception in the future. As the example shown in Fig.28, with the co-visual feature association and relative pose of wide-baseline cameras, we will further explore the remote depth estimation for the map. The remote mapping of vast obstacles in advance will significantly support pre-planning trajectories in unknown urban scenarios. We hope the proposed dual-channel and Rel-MSCKF algorithms could inspire more interesting multi-robot collaborative perception research.

\balance


\bibliographystyle{IEEEtran}
\bibliography{refv3}


\vfill

\end{document}